\documentclass[twoside,11pt]{article}

\usepackage{jmlr2e}

\usepackage{amssymb,amsmath,comment}
\usepackage{amsfonts,mathrsfs}
\usepackage{color}
\usepackage{bbm}
\usepackage{booktabs,float}
\usepackage{graphicx}
\usepackage{enumerate}
\newcommand{\cmmnt}[1]{} 

\makeatletter
\renewcommand*{\thanks}[1]{%
  \footnotemark
  \protected@xdef\@thanks{\@thanks
    \protect\footnotetext[\arabic{footnote}]{#1}}%
}
\makeatother

\ShortHeadings{Excess Risk for Classification with CNNs}{Shen, Jiao, Lin and Huang}
\firstpageno{1}

\begin{document}

\title{Non-asymptotic Excess Risk Bounds for Classification with Deep Convolutional Neural Networks}

\author{\name Guohao Shen\thanks{Guohao Shen and Yuling Jiao contributed equally to this work} \email ghshen@link.cuhk.edu.hk \\
       \addr Department of Statistics\\
       The Chinese University of Hong Kong\\
        Hong Kong, China
       \AND
       \name Yuling Jiao$^*$ 
       \email yulingjiaomath@whu.edu.cn \\
       \addr School of Mathematics and Statistics\\
				Wuhan University\\
Wuhan, Hubei, China 430072
      \AND
\name Yuanyuan Lin \email ylin@sta.cuhk.edu.hk \\
 \addr Department of Statistics\\
The Chinese University of Hong Kong \\
      Hong Kong, China
       \AND
\name Jian Huang \email jian-huang@uiowa.edu\\
\addr       Department of Statistics and Actuarial Science \\
				University of Iowa\\
Iowa City, Iowa 52242,  USA
       }


\maketitle

\vspace{1 cm}
\begin{abstract}
In this paper,  we consider the problem of binary classification with a class of general deep convolutional neural networks, which includes fully-connected neural networks and fully convolutional neural networks as special cases. We establish non-asymptotic excess risk bounds for a class of convex surrogate losses and  target functions with different modulus of continuity. An important feature of our results is that we clearly define the prefactors of the risk bounds in terms of the input data dimension and other model parameters and show that they depend polynomially on the dimensionality in some important models. We also show that the classification methods with CNNs can circumvent the curse of dimensionality if the input data is supported on an approximate low-dimensional manifold.  To establish these results, we derive an upper bound for the covering number for the class of general convolutional neural networks with a bias term in each convolutional layer, and derive new results on the approximation power of CNNs for any uniformly-continuous target functions. These results provide further insights into the complexity and the approximation power of general convolutional neural networks, which are of independent interest and may have other applications. Finally, we apply our general results to analyze the non-asymptotic excess risk bounds for four widely used methods with different loss functions using CNNs, including the least squares, the logistic, the exponential and the SVM hinge losses.\footnote{This version submitted to arXiv on May 01, 2021}
\end{abstract}


\section{Introduction}
\label{intro}
Consider a binary classification problem with a predictor $X\in\mathcal{X}\subset\mathbb{R}^d $ and
its binary label $Y\in
\{1,-1\}$. We are interested in learning a deterministic function $h:\mathcal{X}\to\{1,-1\}$ from a certain class of measurable functions $\mathcal{H}$.
The class  $\mathcal{H}$ is usually referred to as a hypothesis set and any function
$h \in \mathcal{H}$  a classifier. A classifier $h$ can be used for predicting the label of a new observation.
Let the joint distribution of $(X, Y)$ be denoted by $\mathbb{P}$.
The main goal of classification is to find a classifier that minimizes the misclassification error or the 0-1 risk:
\begin{equation}
	R(h) :=\mathbb{E}\big\{\mathbbm{1}(h(X)\not=Y)\big\}=\mathbb{P}\big\{h(X)\not= Y\big\}, h \in \mathcal{H},
\end{equation}
where 
$\mathbbm{1}(E)$ is the indicator function of the set $E$ that is $1$ if event $E$ occurs and $0$ otherwise.
Let the measurable function $h_0:\mathcal{X}\to\{1,-1\}$ be the misclassification risk minimizer at the population level,  i.e.,
\begin{equation}\label{bayes}
	h_0:=\arg\min_{h \in  \mathcal{H} } R (h).
\end{equation}
For any $h\in \mathcal{H}$, the excess risk of $h$ is the difference $R(h)-R(h_0)$ between the misclassification error of $h$ and $h_0$.

Since the probability measure $\mathbb{P}$ is unknown in practice, the classifier $h$ will be learned based on a random sample  $S=\{(X_i,Y_i)\}_{i=1}^n$ from $\mathbb{P}$, where $n$ is the sample size.
By minimizing the empirical excess risk over $\mathcal{H}$,  the empirical risk minimizer (ERM) is defined by
\begin{equation}\label{erm}
		\hat{h}_n\in\arg\min_{h\in\mathcal{H}} \frac{1}{n}\sum_{i=1}^n\mathbbm{1}(h(X_i)\not=Y_i).
\end{equation}	
However, the empirical risk function based on the $0$-$1$ loss is non-continuous and non-convex in its argument, thus the minimization is typically NP-hard and computational intractable.

Rather than minimizing the non-smooth 0-1 loss, many popular methods adopt a proper convex loss function to train classifiers with computational efficiency that can be done in polynomial
time, e.g., logistic regression  \citep{cox1958regression,  kleinbaum2002logistic}, support vector machine (SVM)  \citep{boser92atraining, cortes1995support}, and boosting \citep{freund1996experiments}.
Moreover, proper (classification calibrated) surrogate convex loss functions  have been shown to be consistent with the 0-1 loss function by \citet{zhang2004statistical} and \citet{bartlett2006convexity},  in the sense that the minimizer of proper surrogate convex loss can also produce a minimizer of the generic 0-1 loss for binary classification problems.
\citet{nock2008efficient} and \citet{masnadi2008design} gave convincing arguments that the choice of surrogate loss function for the classification problem depends on the joint distribution of $(X, Y)$.	 Further developments on the surrogate convex loss methods were reported by \citet{ben2012minimizing}, \citet{masnadi2015view} and \cite{telgarsky2015convex}, among others.
A critical issue that affects performance in classification is what hypothesis set $\mathcal{H}$ to choose  \citep{mohri2018foundations}. Popular choices for the hypothesis set have included those based on reproducing kernels \citep{svm-intro1999, scholkopf2018learning} and tree structured models \citep{cart1984, statlearning2001}.

With the rapid developments in deep learning, various classification methods with hypothesis sets
specified through deep neural networks 
have achieved remarkable successes in a variety of machine learning tasks
\citep{
lecun2015deep, schmidhuber2015}.
In particular,
\textit{convolutional neural networks} (CNN) have demonstrated outstanding performance
in many applications, including computer vision \citep{krizhevsky2012imagenet}, natural language processing \citep{wu2016google}, and sequence analysis in bioinformatics \citep{alipanahi2015predicting,zhou2015predicting}.
However, to the best of our knowledge, there have been no systematic studies on the properties of
various classification methods using CNNs in terms of the non-asymptotic bounds for the excess risks.

The excess risk of an empirical risk minimizer over CNNs is controlled by two types of error: estimation error and approximation error. \cite{zhang2004statistical}, \cite{bartlett2006convexity} and \cite{mohri2018foundations} established upper bounds on the estimation error for a class of convex surrogate losses.
In recent years, several studies have
provided insights into
the empirical successes of CNNs from a theoretical standpoint in terms of the capacity of CNNs and their approximation power.
\cite{bartlett2017spectrally} and \cite{golowich2018size} derived upper bounds for the covering number and Rademacher complexity of fully connected neural networks without the bias term. Based on the same technique, \cite{lin2019generalization} proved an upper bound of general convolutional neural networks which allows arbitrarily ordered fully connected layers and convolutional layers without bias. A similar result was proved  by \cite{zhou2018understanding} for the fully convolutional networks with the last layer being fully connected.  \cite{li2018tighter} studied generalization error bounds of CNNs composed
of orthogonal filters.  \cite{arora2018stronger} also gave complexity bounds for deep networks via a compression approach. We provide a more detailed discussion of the related works in Section \ref{related} below.

\subsection{Our contributions}
\label{contributions}
 In this paper, we consider the problem of binary
 classification with the hypothesis set $\mathcal{H}$ specified using general deep convolutional neural networks, which includes fully-connected neural networks and fully convolutional neural networks as special cases.  Our main contributions are as follows.

 \begin{enumerate}[(i)]
  \setlength\itemsep{-0.03 cm}
 \item
 We establish non-asymptotic excess risk bounds for a class of classification methods using
CNNs. Our results are derived for a general class of convex surrogate losses and target functions with different modulus of continuity. They are applicable to many widely used classification methods in applications, including those with the least squares, the logistic (cross-entropy), the exponential
and the SVM hinge loss functions.
\item
An important feature of our risk bounds is that we clearly define the prefactors in terms of the input data dimension $d$ and other model parameters.  Our risk bounds
 significantly improve over the currently available ones in the sense that
 they depend polynomially
 on the dimensionality $d$ in some important models including the logistic regression and SVM.
 This is  important since the prefactors can dominate the error bounds
 in high-dimensional problems.
  However, the prefactor of the excess risk bound with the exponential loss
 depends on $d$ exponentially. We provide an explanation for the differences in the characteristics of
 the prefactors in terms of the Lipschitz constants of these loss functions.

 \item We show that the classification methods with CNNs can circumvent the curse of dimensionality if the input data is supported on a neighborhood of a low-dimensional manifold embedded in the ambient space $\mathbb{R}^d$.	
 \item
We establish an upper bound for the covering number for the class of general convolutional neural networks with a bias term in each convolutional layer, and derive new results on the approximation power of CNNs for any uniformly-continuous target functions. These results provide further insights into the complexity and the approximation power of general convolutional neural networks, which are of independent interest and may have applications in other machine learning tasks involving CNNs.

\item
We apply our general results to analyze the non-asymptotic excess risk bounds for four widely used
classification methods with different losses using CNNs, including the least squares, the logistic (cross-entrop), the exponential,  and the SVM hinge loss functions.
\end{enumerate}

The rest of the paper is organized as follows.   In section \ref{sec2} we describe the setup of the problem and the basic framework of convex surrogate loss functions for classification.  We also introduce
a general class of convolutional neural networks to be used in estimating the discriminant function. 		
In section \ref{sec3} we present a basic inequality for the excess risk in terms of the estimation error
and the approximation errors. 	
An 
upper bound for the covering numbers of CNNs with bias terms in each of its layers is derived for analyzing these errors.  In section \ref{sec4}
we establish the consistency and an explicit bound for the excess risk of empirical risk minimizer based on the surrogate loss and its induced classifier under mild conditions, with and without the approximate low-dimensional manifold assumption on the distribution of the input data. We also derive an upper bound on the approximation error of CNNs for any uniformly-continuous target functions under the approximate low-dimensional manifold assumption.
Several important examples, including the logistic,
the SVM  and the exponential losses are provided to illustrate our general convergence results in Section \ref{examples}. In Section \ref{related} we discuss related works and comment on the differences of our work with the existing ones.
Discussions and closing remarks are given in section \ref{conclusion}. Technical proofs are deferred to Appendix.

\section{ 
Preliminaries}
\label{sec2}
 In this section, we present the basic setup of
 the binary classification problem and define the excess risk. We also introduce the convex surrogate loss function and describe the structure of convolutional neural networks to be used in the estimation of the discriminant function.

\subsection{Misclassification error}
\label{erm1}

The performance of a classification method is determined by the following three factors: (i) the
joint distribution of $X$ and $Y$, which determines how well $X$ can predict $Y$; (ii) the loss function
$\ell$ and the hypothesis set $\mathcal{H}$; and (iii) the prediction rule.
There are 
two possible scenarios for the relationship between $X$ and $Y$: a deterministic relationship and a
stochastic one.
In the deterministic scenario, it is assumed that $Y=f_0(X)$, i.e., $Y$ is 
completely determined by $X$ through a deterministic function $f_0:\mathcal{X}\to\{1,-1\}$.
While in the stochastic scenario, the label $Y$ cannot be solely decided by $X$, and the relationship between $Y$ and $X$ is described by the conditional probability
\begin{equation}
\label{eta}
 \eta(x):=\mathbb{P}(Y=1\mid X=x), \ x \in \mathcal{X}.
\end{equation}  
As will be seen below, this conditional probability function plays a central role in our analysis.
The stochastic scenario is more realistic and general. In this case, one cannot correctly predict the label of $X$ with certainty 
unless $\eta$ only takes 1 or 0 on $\mathcal{X}$, for which the stochastic scenario reduces to the deterministic one.

To elaborate, for any $x\in\mathcal{X}$ with $\eta(x)\in(0,1)$, the label $Y$ for a given $X=x$ could be $1$ or $-1$. Let ${\rm sign}(\cdot)$ be the sign function, i.e.,  ${\rm sign}(x)=1$ if $x\geq0$ and ${\rm sign}(x)=-1$ otherwise. Then, the best prediction for $Y$ for a given $X=x$ that achieves minimal prediction error is the {\it Bayes classifier}
$$h_0(x)={\rm sign}(\eta(x)-1/2)=\arg\max_{y\in\{1,-1\}}\mathbb{P}(Y=y\mid X=x).$$
However,  one may still have $\min\{\eta(x),1-\eta(x)\}$ chance of
 making a mistake in the prediction. When  $\eta(x)=1/2$, the predicted label being $1$ or $-1$ makes no difference in terms of the  probability of making a mistake.  Correspondingly, for each $x\in\mathcal{X}$,  we define the noise function ${\rm noise}(x):=\min\{\eta(x),1-\eta(x)\}$.
The noise is a characteristic of the learning task indicative of its level of difficulty, and the Bayes classifier achieves the minimal possible risk $R_0\equiv R(h_0)=\mathbb{E}\{{\rm noise}(X)\}$. The risk of the Bayes classifier is often viewed as a benchmark.  Our goal is to
 to find an $h\in\mathcal{H}$  that minimizes $R(h)-R_0$.
	
However, the minimization problem (\ref{erm}) is computational intractable, as the empirical risk function is non-continuous and non-convex in its argument.   Therefore, it is difficult to solve this empirical minimization problem in practice.
One of the prominent methods to alleviate this problem is to work with a surrogate for the $0$-$1$ loss, so that the associated optimization problem can be solved efficiently
\citep{lin2004, lugosi2004, zhang2004statistical, bartlett2006convexity}.

\subsection{Convex surrogate loss functions}\label{sec2.1}
We next give a brief review of the convex surrogate loss functions and discuss the connection between the excess risk with respect to the $\phi$-loss and that of 0-1 loss \citep{zhang2004statistical, bartlett2006convexity}.
	
Let $\phi$ be a given convex univariate function $\phi:\mathbb{R}\to[0,\infty)$.
	 Instead of minimizing the excess risk $R$ over $\mathcal{H}$,  we consider minimizing
	the risk with respect to the loss $\phi$ ($\phi$-risk)
	\begin{align*}
	 R^\phi(f) :=\mathbb{E}\{\phi(Yf(X))\}
	 \end{align*}
	 over a certain class of functions $\mathcal{F}$, where $\phi:\mathbb{R}\to[0,\infty)$ is some generic loss function.
For the special case when $\mathcal{H}=\{h:h(x)={\rm sign}(f(x)), f\in\mathcal{F}\}$ and $\phi(\cdot)$ is a step function, i.e.,  $\phi(x)=1$ if $x<0$ and $\phi(x)=0$ if $x\geq0$,  then to minimize ${R}$ over $\mathcal{H}$, it suffices to minimize $R^\phi$ over $\mathcal{F}$.  Given $S=\{(X_i,Y_i)\}_{i=1}^n$,
let  $R^\phi_n (f) :=\sum_{i=1}^{n}\phi(Y_if(X_i))/n$ be the empirical risk of $f$  w.r.t the $\phi$-loss,  and define the empirical risk estimator (ERM) as
	\begin{equation}\label{germ}
		\hat{f}^\phi_n\in\arg\min_{f\in\mathcal{F}} R^\phi_n (f).
	\end{equation}
Based on the ERM $\hat{f}^\phi_n$, a classification rule or a classifier  $\hat{h}_n^\phi(x) :={\rm sign}(\hat{f}^\phi_n(x))$ for $x\in\mathcal{X}$ can be induced which aims at minimizing the 0-1 risk in (\ref{erm}).
As shown in \cite{zhang2004statistical} and \cite{bartlett2006convexity}, for a properly chosen $\phi$, $\hat{f}^\phi_n$ can indeed help reduce the excess risk $R(\hat{h}^\phi_n)-R_0$.
More precisely,
	 let $R^\phi_0 :=\inf_{f {\ \rm measurable}}R^\phi(f)$, then for a proper $\phi$, we have $$\psi(R(\hat{h}^\phi_n)-R_0)\leq  R^\phi(\hat{f}_n^\phi)-R^\phi_0,$$
	where $\psi:[-1,1]\to[0,\infty)$ is a nonnegative continuous function,  invertible on $[0,1]$, and achieves its minimum at $0$ with $\psi(0)=0$.
	A wide variety of popular classification methods are based on this tactic. For instance,
	when $\phi(a)=\log\{1+\exp(-a)\}$,  it is the  logistic regression  \citep{friedman2000};  when  $\phi(a)=\max\{1-a,0\}$, it becomes the
	 SVM \citep{cortes1995support};
when $\phi(a)=\exp(-a)$ is the exponential loss, 	 it is the AdaBoost algorithm \citep{freund1996experiments};
when $\phi(a)=(1-a)^2$,
it is the least squares method for classification, and so on.
	
Define the measurable minimizer of $R^\phi$ as
	\begin{equation}\label{ermphi}
		f^\phi_0=\arg\min_{f {\ \rm measurable}} \mathbb{E}\{\phi(Yf(X))\},
	\end{equation}
   	and the corresponding minimal $\phi$-risk as $R^\phi_0=R^\phi(f^\phi_0)$.
	Clearly, the optimal $f^\phi_0$ depends on the loss function $\phi$. If $\phi$ is not properly chosen, the resulting $f^\phi_0$ can be poor and thus the classifier ${h}_0^\phi(x)={\rm sign}({f}^\phi_0(x))$ based on $f^\phi_0$ can be invalid. To study the basic conditions imposed on $\phi$, so as to produce a valid classifier towards minimizing the 0-1 loss, we need a thorough understanding of
the risk $R^\phi$.
	
Let the conditional $\phi$-risk of $f$ given $X=x$ be denoted by $R_{x}^\phi(f):=\mathbb{E}\{\phi(Yf(X))\mid X=x\}$. 	 Recall that $\eta(x)=\mathbb{P}\{Y=1\mid X=x\}$ as defined in (\ref{eta}).  We have
	$$R_{x}^\phi(f)=\eta(x)\phi(f(x))+\{1-\eta(x)\}\phi(-f(x)), \ x\in\mathcal{X}.
 $$
For a good classifier, if $\eta(x)>1/2$,  it is naturally expected that  $f_0^\phi(x)>0$ and $h_0^\phi(x)={\rm sign}(f^\phi_0(x))=1$ (correct sign);
	thus to encourage $f_0^\phi(x)>0$, we should at least require $\phi(f_0^\phi(x))<\phi(-f_0^\phi(x))$ to minimize the conditional risk $R^\phi_x$. Similarly, if $\eta(x)<1/2$,  we expect that the contrary happens. A rigorous definition for such an ideal  $\phi$ is given in Definition 1 of \cite{bartlett2006convexity}.
	\begin{definition}[Classification-calibrated]
		For $\eta\in[0,1]$ and $a\in\mathbb{R}$, define $H^\phi(\eta,a)=\eta\phi(a)+(1-\eta)\phi(-a)$. Then,
		we say that $\phi$ is {\it classification-calibrated} if for any $\eta\not=1/2$,
		$$\inf_{a\in\mathbb{R}}H^\phi(\eta,a)<\inf_{a\in\mathbb{R}:a(2\eta-1)\leq0}H^\phi(\eta,a).$$
	\end{definition}
	
	With a classification-calibrated loss $\phi$, it is guaranteed that the ``incorrect" label ${\rm sign}(f(X))$
in the sense that it is inconsistent with the Bayes estimator ${\rm sign}(2\eta(X)-1)$) results in a strictly larger loss under $\phi$. It was shown by \cite{bartlett2006convexity} that, the surrogate loss function $\phi$ is 
able to
produce
the optimal Bayes classifier. For ease of reference, we state this result as Lemma \ref{surrogate} below.  	
	\begin{lemma}[Theorem 1 of \cite{bartlett2006convexity}]\label{surrogate}
	For any nonnegative loss function $\phi:\mathbb{R}\to[0,\infty)$, any measurable $f:\mathcal{X}\to\mathbb{R}$, its induced classifier $h_f={\rm sign}(f):\mathcal{X}\to\{\pm1\}$, and any probability measure $\mathbb{P}$ of $(X,Y)$ on $\mathcal{X}\times\{\pm 1\}$,
	$$\psi(R(h_f)-R_0)\leq R^\phi(f)-R^\phi_0,$$
	where $\psi:[-1,1]\to[0,\infty)$ is the Fenchel-Legendre biconjugate of $\tilde{\psi}:[-1,1]\to\mathbb{R}$, and $$\tilde{\psi}(\theta)=\inf_{a\in\mathbb{R}:a(2\eta-1)\leq0}H(\frac{1+\theta}{2})-\inf_{a\in\mathbb{R}}H(\frac{1+\theta}{2}).$$
	Besides, if $\phi$ is classification-calibrated, then for any sequence $\{a_m\}$ in $[0,1]$, $\psi(a_m)\to0$, if and only if $a_m\to0$.
	\end{lemma}
Lemma \ref{surrogate} has some important implications. First, for any nonnegative $\phi$, $\psi$ is simply the functional convex hull (the greatest convex minorant) of $\tilde{\psi}$. Both $\psi$ and $\tilde{\psi}$ are continuous on $[-1,1]$, and $\psi$ is nonnegative that attains its minimum at 0 with $\psi(0)=0$. Second, if the loss $\phi$ is nonnegative and classification-calibrated, then $\psi(\theta)>0$ for all $\theta\in(0,1]$ and $\psi$ is invertible on $[0,1]$. In this case, we have $R(h_f)-R_0\leq \psi^{-1}(R^\phi(f)-R^\phi_0)$. Third, if $\phi$ is nonnegative convex and classification-calibrated, then $\psi(\theta)=\phi(0)-\inf_{a\in\mathbb{R}}H^\phi((1+\theta)/2,a)$, which gives an easy way to compute the function $\psi$.
	
Next,  we present several examples of $\phi$, and the corresponding $f_0^\phi$, $R^\phi(f_0^\phi)$, $\psi$ and its inverse on $[0,1]$ in Table \ref{tab:1}.

\begin{table}[H]
	\caption{Minimizer and minimal conditional risk under different loss functions $\phi$.}
	\label{tab:1}
	\centering
	\resizebox{\textwidth}{!}{%
	\begin{tabular}{@{}lllll@{}}
		\toprule
		& Least squares     & SVM        &Exponential   		&Logistic          \\ \midrule
		$\phi(a)$           & $(1-a)^2$  & $\max\{1-a,0\}$      & $\exp(-a)$                                   & $\log\{1+\exp(-a)\}$                                 \\
		$f^\phi_0(x)$        & $2\eta-1$ & ${\rm sign}(2\eta-1)$ & $\frac{1}{2}\log(\frac{\eta}{1-\eta})$ & $\log(\frac{\eta}{1-\eta})$                    \\
		$R^\phi_x(f^\phi_0)$ & $4\eta(1-\eta)$    & $1-\vert 2\eta-1\vert$                              & $2\sqrt{\eta(1-\eta)}$               & $-\eta\log(\eta)-(1-\eta)\log(1-\eta)$ \\
		$\psi(\theta)$		& $\theta^2$ & $\vert\theta\vert$ & $1-\sqrt{1-\theta^2}$ &$\frac{1+\theta}{2}\log(\frac{1+\theta}{1-\theta})+\log(1-\theta)$ \\ 	
		$\psi^{-1}(\theta)$ & $\sqrt{\theta}$ & $\vert \theta\vert$ & $\sqrt{1-(1-\theta)^2}$ & *** \\ \bottomrule
	\end{tabular}
}
{\begin{flushleft}
	Note: $\eta(x)$ is written as $\eta$ for notational  simplicity. ``***" stands for no explicit form.
	\end{flushleft}  }
\end{table}

\subsection{Convolutional neural networks} \label{sec2.2}
As indicated by their name, CNNs employ a mathematical operation called convolution. Convolutional networks are a specialized type of structured sparse feedforward neural network (FNN) that use convolution in place of general matrix multiplication in at least one of their layers
 \citep{goodfellow2016deep}. There are different formulations of CNNs in the literature \citep{bao2014approximation,zhou2018understanding,oono2019approximation,lin2019generalization,zhou2020theory,zhou2020universality}.
 In this paper, we consider a generic formulation of CNNs  with  bias vectors in the convolutional layers,
 	which is a generalization of that defined in \cite{lin2019generalization} and covers the multi-layer perceptions (MLP), fully convolutional neural networks (FCNNs) and general convolutional neural networks (CNNs), i.e neural networks with arbitrarily ordered and mixed layers (each layer can be fully convolutional layer or fully-connected layer).

For a general neural network, let $L$ denote the number of layers and  $(\sigma_1,\ldots,\sigma_L)$ denote the activation functions, such as the rectified linear unit (ReLU) and max pooling function. In this paper, it is assumed that $\sigma_i(0)=0$ and $\sigma_i$ is $\rho_i$-Lipschitz for $i=1,\ldots, L$. Besides,  let $d_i$ denote the width (the number of neurons or computational units) of the $i$-th layer, $i=1,\ldots,L$,  and $\mathcal{W}=\max\{d_1,\ldots,d_L\}$ denote the maximum width among layers. In our  present classification problem, $d_1=d$ (the dimension of the input $X$) and $d_L=1$ (the dimension of the output $Y$).
	
Next, we describe the architectures of  MLPs, FCNNs and CNNs.
	\begin{itemize}
		\item The architecture of a MLP $f_{MLP}:\mathbb{R}^d\to\mathbb{R}$ can be expressed as a composition of a series of linear transformations: for any $x\in\mathbb{R}^d$,
		\begin{equation}\label{MLP}		f_{MLP}(x)=\sigma_L(W_{L}\sigma_{L-1}(\cdots\sigma_2(W_2\sigma_1(W_1x+b_1)+b_2)\cdots)+b_L),
		\end{equation}
	 	where $W_i\in\mathbb{R}^{d_{i+1}\times d_i}$  is a fully connected weight matrix, 		
		 and $b_i\in\mathbb{R}^{d_{i+1}}$ is a bias vector in the $i$-th linear transformation. Usually, the activation functions $(\sigma_1,\ldots,\sigma_L)$ are the ReLU activation, where $\mbox{ReLU}(x)=\max\{0,x\}$ for any $x\in\mathbb{R}$ (defined for each component of $x$ if $x$ is a vector).
		
		\item The architecture of a FCNN $f_{FCNN}:\mathbb{R}^d\to\mathbb{R}$ is similar to that of MLP, but with structured sparse weight matrix. For any $x\in\mathbb{R}^d$,
		\begin{equation}\label{FCNN} f_{FCNN}(x)=\sigma_L(W^c_{L}\sigma_{L-1}(\cdots\sigma_2(W^c_2\sigma_1(W^c_1x+b^c_1)+b^c_2)\cdots)+b^c_L),
		\end{equation}
		where  $b^c_i\in\mathbb{R}^{d_{i+1}}$ is the bias vector, and $W^c_i\in\mathbb{R}^{d_{i+1}\times d_i}$ is the structured weight matrix induced by the convolutional weights $C_i\in\mathbb{R}^{r_i\times s_i}$ which contains $r_i$ convolutional filters, each of which has dimension (or size) $s_i$, $i=1,\ldots,L$.
For FCNN, the activation functions are taken to be the ReLU or  the max pooling function. We 
elucidate what the max pooling function is and how the convolutional weight $C_i$  generates $W^c_i$ later in this subsection.
		
\item The formulation of a CNN $f_{CNN}:\mathbb{R}^d\to\mathbb{R}$ is essentially a combination of the architectures of MLP and FCNN,
		\begin{equation}\label{CNN}
		 f_{CNN}= \sigma_{L}\circ A_L\cdots\circ\sigma_2\circ A_2\circ\sigma_1\circ A_1,
		\end{equation}
		where $\circ$ denotes the functional composition and $A_i$ is a linear operation, $i=1,\ldots,L$. The form of $A_i$ is either $$A_i(x)=W_ix+b_i,$$ where $W_i\in\mathbb{R}^{d_{i+1}\times d_i}$ and $b_i\in\mathbb{R}^{d_{i+1}}$,  or $$A_i=W^c_ix+b^c_i, $$ where $W^c_i\in\mathbb{R}^{d_{i+1}\times d_i}$ is the structured sparse weight matrix induced by the convolutional weights $C_i\in\mathbb{R}^{r_i\times s_i}$, and $b^c_i\in\mathbb{R}^{d_{i+1}}$ is  bias vector. Basically, each layer of a CNN is either a fully connected layer or a convolutional layer.
	\end{itemize}

	Now, we illustrate how structured weight matrices are generated by convolution operations for convolutional layers. Consider a single convolutional layer, let $C=(c_1,\ldots,c_r)\in\mathbb{R}^{r\times s}$ denote the $r$ convolutional filters with the same dimension $s$, and $X\in\mathbb{R}^d$ be an input vector. For each filter $c_i$, the convolutional operation performs $M$-times inner product on $c_i$ and an $s$-dimensional subvector of the $d$-dimensional covariate $X$. For the $m$-th operation, let
	$S_m\in\mathbb{N}_+^s$ be the index set consisting of the indices of the $s$-dimensional subvector $X_{S_m}$ of $X$. Then, the $m$-th linear combination for the filter $c_i$ is
	$$c_i^\top X_{S_m}=c_{i,S_m}^\top X,$$
	where $c_{i,S_m}\in\mathbb{R}^d$ is expanded $c_i$, i.e. elements of $c_{i,S_m}$ are re-arranged to the places indexed by $S_m$ while filling the other places with zeros.
	For a better understanding, an example as in \cite{lin2019generalization} is illustrated here. Consider a 2-D convolutional filter $c\in\mathbb{R}^{2\times2}$ (with dimension or size 4) and a 2-D input (2 channels) $X\in\mathbb{R}^{3\times4}$ as follows:
	$$c=
	\left[
	\begin{matrix}
	c_{1,1} & c_{1,2} \\
	c_{2,1} & c_{2,2}
	\end{matrix}
	\right], {\ \rm and\ }
	X=\left[
	\begin{matrix}
		X_{1,1} & X_{1,2} & X_{1,3} & X_{1,4} \\
		X_{2,1} & X_{2,2} & X_{2,3} & X_{2,4} \\
		X_{3,1} & X_{3,2} & X_{3,3} & X_{3,4}
	\end{matrix}
	\right].
	$$
	A standard convolution operation with stride (or step size) one performs $m=6$ times linear combination orderly. To help understand the operations here, imagine $c$ as a pad sliding on the surface of another pad $X$ each step for one time, from the top-left (when $c_{1,1}$ aligns with $X_{1,1}$ and $c_{2,2}$ aligns with $X_{2,2}$) to the bottom-right (when $c_{1,1}$ aligns with $X_{2,3}$ and $c_{2,2}$ aligns with $X_{3,4}$). For each slide, we move $c$ by one step (one grid of matrix), match the respective elements of $c$ and $X$, calculate their inner product, and save the result into a proper element of the resulting matrix. Finally, the resulting matrix is
	$$  c*X \equiv
	\left[
\begin{matrix}
	\sum_{i=1,j=1}^{2} c_{i,j}X_{i,j}\ \ & \sum_{i=1,j=1}^{2} c_{i,j}X_{i,j+1}\ \ & \sum_{i=1,j=1}^{2} c_{i,j}X_{i,j+2}  \\
	\sum_{i=1,j=1}^{2} c_{i,j}X_{i+1,j} & \sum_{i=1,j=1}^{2} c_{i,j}X_{i+1,j+1} & \sum_{i=1,j=1}^{2} c_{i,j}X_{i+1,j+2}  \\
\end{matrix}
\right].
$$
Here,  $c*X $ denotes the convolution of matrix $c$ and $X$.

Let $X^\prime=(X_{1,1} , X_{1,2} , X_{1,3} , X_{1,4}, X_{2,1} , X_{2,2} , X_{2,3} , X_{2,4} ,X_{3,1} , X_{3,2} , X_{3,3} , X_{3,4})^\top$ be the 1-D vector of reshaped $X$. Then, the induced weight matrix $W^c$ of $c$ is
\setcounter{MaxMatrixCols}{12}
\begin{equation*}
	W^c=
	\left[
\begin{matrix}
 c_{1,1} & c_{1,2} & 0 & 0 & c_{2,1} & c_{2,2} & 0 & 0 & 0 & 0 & 0 & 0  \\
0 & c_{1,1} & c_{1,2} & 0 & 0 & c_{2,1} & c_{2,2} & 0 & 0 & 0& 0 & 0 \\
0 & 0 & c_{1,1} & c_{1,2} & 0 & 0 & c_{2,1} & c_{2,2} & 0 & 0 & 0 & 0 \\
0 & 0 & 0 & 0 & c_{1,1} & c_{1,2} & 0 & 0 & c_{2,1} & c_{2,2}& 0 & 0 \\
0 & 0 & 0 & 0 & 0 & c_{1,1} & c_{1,2} & 0 & 0 & c_{2,1}& c_{2,2} & 0 \\
0 & 0 & 0 & 0 & 0 & 0 & c_{1,1} & c_{1,2} & 0 & 0 & c_{2,1} & c_{2,2} \\
\end{matrix}
	\right].
\end{equation*}
Note that $W^cX^\prime\in\mathbb{R}^6$ is the 1-D vector of the reshaped $c*X$. With the above notation, the induced weight matrix $W^c$ of a convolutional matrix $C=(c_1,\ldots,c_r)$ with $r$ filters, each of which performs $m$ operations, can be written as
$$W^c=(c_{1,S_1},\ldots,c_{1,S_m},c_{2,S_1},\ldots,c_{2,S_m},\ldots,c_{r,S_1},\ldots,c_{r,S_m})^\top\in  \mathbb{R}^{d_{out}\times d_{in},   }$$
where $d_{in}$ is the dimension of the input of this layer, $d_{out}$ is the dimension of the output of the layer and $d_{out}=rm$. For 3-D or tensor data with channels more than 2, the filters can be designed to accommodate them without further difficulty, and the induced weight matrix can be formulated in a similar fashion.

Max-pooling function
is a mapping from a finite-dimensional vector space (possibly arranged as matrices or tensors) to another one. Given a vector $U=(u_1,\ldots,u_{d_{in}})^\top\in\mathbb{R}^{d_{in}}$, the max-pooling operator iterates
over a collection of sets of indices $\mathcal{I}$ and outputs $V=(v_1,\ldots,v_{d_{out}})^\top\in\mathbb{R}^{d_{out}}$, where the cardinality of $\mathcal{I}$ equals to the dimension of the output $d_{out}$. For each element $I_i\in\mathcal{I}$, the $i$-th component of the output $V$ is the maximal entry of the input $U$ over index $I_i$, i.e.
$V_i:=\max_{j\in I_i}U_j.$

As shown in Lemma A.2 in \cite{bartlett2017spectrally}, the Lipschitz constant of the max-pooling operators depends on the number of times each coordinate is accessed across elements of $\mathcal{I}$. To be precise, if each coordinate of the input  appears in at most $m$ elements of the collection $\mathcal{I}$, then the max-pooling operator is $m^{1/p}$-Lipschitz w.r.t the $L_p$-norm $ \Vert\cdot\Vert_p$  for any $p\geq1$. The max-pooling operator is 1-Lipschitz whenever $\mathcal{I}$ forms a partition. In practice, for computer vision tasks, the number of times $m$ in the max-pooling operator is often a small constant.

	\section{Excess $\phi$-risk}\label{sec3}
	In this section,  we present  a basic inequality for the excess $\phi$-risk and focus the discussions on the stochastic error and approximation error.

	\subsection{Excess risk decomposition}
	\label{sec3.1}
	With a nonnegative, convex and classification-calibrated surrogate loss function $\phi$, it is no longer necessary to minimize the excess 0-1 risk $R(h_f)-R_0$, instead we turn to minimize the excess $\phi$-risk $R^\phi(f)-R^\phi_0$ and rest assured that
	$$R(h_f)-R_0\leq \psi^{-1}(R^\phi(f)-R^\phi_0).$$
	From now on, we assume  that $\phi$ is nonnegative, convex and classification-calibrated, and focus our study on the excess $\phi$-risk  $R^\phi(f)-R^\phi_0$.
	
To begin with,  for any estimator $f$ belonging to a certain class of functions $\mathcal{F}$, 	
	its excess $\phi$-risk can be decomposed as \citep{mohri2018foundations}
	$$R^\phi(f)-R^\phi(f^\phi_0)=\left\{ R^\phi(f)-\inf_{f\in\mathcal{F}}R^\phi(f)\right\}+ \left\{\inf_{f\in\mathcal{F}}R^\phi(f)- R^\phi(f^\phi_0) \right\},$$
	where $f^\phi_0$ is defined in (\ref{ermphi}). The first term of the right hand side is the {\it estimation error}, and the second term is  the {\it approximation error}.
	The estimation error depends on the estimator $f$, which measures the difference of the error of $f$ and the best one in $\mathcal{F}$. The approximation error only depends on the function class $\mathcal{F}$, which measures how well the function $f_0^\phi$ can be approximated using $\mathcal{F}$
	with respect to the loss $\phi$. For a complex and rich function class $\mathcal{F}$, the approximation error is likely to be small, while at the same time the estimation error is likely to be large.
	
For the classification problem under investigation,  $f$ is the ERM  $\hat{f}^\phi_n$ defined in (\ref{germ}).
A general upper bound on the excess risk of $\hat f_n^\phi$  is given below.
	
\begin{lemma}\label{lemma2}
For any  loss  $\phi$ and any random sample $S=\{(X_i,Y_i)\}_{i=1}^n$,  the excess $\phi$-risk of the ERM  $\hat{f}^\phi_n$ satisfies
		\begin{align}
			R^\phi(\hat{f}^\phi_n)-R^\phi(f^\phi_0)\leq&  2\sup_{f\in\mathcal{F}}\vert R^\phi(f)-R^\phi_n(f)\vert+\inf_{f\in\mathcal{F}}R^\phi(f) -R^\phi(f^\phi_0).
		\end{align}
\end{lemma}	
	
  The excess risk of  $\hat f_n^\phi$ is bounded above by the sum of two terms: the stochastic error $2\sup_{f\in\mathcal{F}}\vert R^\phi(f)-R^\phi_n(f)\vert$ and the approximation error $\inf_{f\in\mathcal{F}}R^\phi(f)-R^\phi(f^\phi_0)$. Notice that the upper bound no longer depend on the ERM, but the function class $\mathcal{F}$, the loss function $\phi$ and the random sample $S$. The first term $2\sup_{f\in\mathcal{F}}\vert R^\phi(f)-R^\phi_n(f)\vert$ is closely related to the complexity of $\mathcal{F}$ and the second term $\inf_{f\in\mathcal{F}}R^\phi(f)-R^\phi(f^\phi_0)$ measures the approximation power of the function class $\mathcal{F}$ for $f_0^\phi$.
	
Throughout the paper, we choose $\mathcal{F}$ to be a function class $f:\mathcal{X}\to\mathbb{R}$ consisting of convolutional neural networks, denoted by $\mathcal{F}_{CNN}$.
To give a  mathematical definition of the function class $\mathcal{F}_{CNN}$,  more notations are needed.
 Let $(\sigma_1,\ldots,\sigma_{L})$ be fixed Lipschitz mappings (each $\sigma_i$ is $\rho_i$-Lipschitz). Let $\mathcal{L}_F$ be a subset of indices $\{1,\ldots,L\}$ and $\mathcal{L}_C=\{1,\ldots,L\}\backslash\mathcal{L}_F$ be its complement, and let $A=(A_1,\ldots,A_L)$ be $L$ linear operators with their respective input dimension $(d_1,\ldots,d_L)$. Here,  for  $l\in\mathcal{L}_F$, $A_l(X)=W_lX+b_l$  with fully connected weights $W_l\in\mathbb{R}^{d_l\times d_{l+1}}$ and bias $b_l\in\mathbb{R}^{d_{l+1}}$;   for $l\in\mathcal{L}_C$, $A_l(X)=W^c_lX+b^c_l$ with  $W^c_l\in\mathbb{R}^{d_l\times d_{l+1}}$ induced by convolutional weights $C_l\in\mathbb{R}^{r_i\times s_i}$ and bias $b^c_l\in\mathbb{R}^{d_{l+1}}$. Let $m_l$ denote the number of operations performed by each of the $r_l$ filters in convolutional layer $l\in\mathcal{L}_C$.
 Let $(a_1,\ldots,a_L)$ be positive numbers uniformly bounded by $B_a>0$ and let $(\mathcal{A}_1,\ldots,\mathcal{A}_{L})$ be sets with $\mathcal{A}_l=\{A_l:\Vert W_l\Vert_F + \Vert b_l\Vert_F\leq a_l\}$ for $l\in\mathcal{L}_F$ and $\mathcal{A}_l=\{ A_l:\Vert C_l\Vert_F+\Vert b^c_l\Vert_F\leq a_l\}$ for $l\in\mathcal{L}_C$,
 where $ \Vert \cdot\Vert_F $ is the Frobenius norm.
 For a general function class $\mathcal{F}$ and any $f_1,f_2\in\mathcal{F}$,   define the $\Vert\cdot\Vert_\infty$ metric on $\mathcal{F}$ by $\sup_{X\in\mathcal{X}}\Vert f_1(X)-f_2(X)\Vert_\infty$.
 Then,  the function class consisting of CNNs is defined as
	\begin{equation}\label{CNNs}
		\mathcal{F}_{CNN}=\{f:\mathcal{X}\to\mathbb{R}^{d_L}:f= \sigma_{L}\circ A_L\cdots\circ\sigma_1\circ A_1, A_l\in\mathcal{A}_l {\ \rm for\ } l=1,\ldots,L\}.
	\end{equation}
	 Let $\mathcal{B}:=\sup_{f\in\mathcal{F}_{CNN}}\Vert f\Vert_\infty$ denote the bound of functions in $\mathcal{F}_{CNN}$,  $\mathcal{S}=\sum_{l\in\mathcal{L}_F}(d_{l-1}+1)\times d_{l}+\sum_{l\in\mathcal{L}_C}(s_i+m_i)\times r_{l}$ be the number of parameters for networks in $\mathcal{F}_{CNN}$ and let $\mathcal{W}=\max\{d_1,\ldots,d_L\}$ be the maximum width among layers.
	
	 For a given sample $S=\{(X_i,Y_i)\}_{i=1}^n$,	with a slight abuse of notation, we still denote the  ERM  over $\mathcal{F}_{CNN}$ w.r.t loss $\phi$ by
	\begin{equation}\label{erm-cnn}
		\hat{f}^\phi_n\in\arg\min_{f\in\mathcal{F}_{CNN}} \frac{1}{n}\sum_{i=1}^n\phi(Yf(X_i)).
	\end{equation}	

\subsection{Estimation error}
\label{sec3.2}
In this subsection, we focus on the estimation error of ERM based on
CNNs and give an upper bound on the estimation error. To this end, we first establish an
upper bound for the covering numbers of CNNs with a bias term in each of its layers.
	
\begin{theorem}[Covering number of CNNs]\label{bound-cvnum-cnn}
		Let $\mathcal{F}_{CNN}$ be the class of functions defined in (\ref{CNNs}). Then, for any $\epsilon>0$,
		$$\log\mathcal{N}(\mathcal{F}_{CNN},\epsilon,\Vert\cdot\Vert_\infty)\leq2 \mathcal{S}\log\Big(1+\frac{L\mathcal{B}}{\epsilon}\Big),$$
		where $\mathcal{N}(\mathcal{F}_{CNN},\epsilon,\Vert\cdot\Vert_\infty)$ denotes the covering number of $\mathcal{F}_{CNN}$ under metric $\Vert\cdot\Vert_\infty$ with radius $\epsilon$.
	\end{theorem}

The covering number in Theorem \ref{sec3.2} is derived by an approach similar to 	
 that in \citet{bartlett2017spectrally}, \citet{golowich2018size} and \citet{lin2019generalization}. In these works, the covering number of a given neural network is upper bounded by terms relevant to the complexity measures depending on the Lipschitz constants of the activation functions and the norms of weight matrices. Similar to \cite{lin2019generalization} but different from many other methods,  our upper bound of the covering number is represented by parameters of the general convolutional neural networks that contain both fully-connected layers and convolutional layers, and the complexity is bounded by the norm of convolutional weights rather than the induced weight matrix. In addition, unlike many existing works, the general convolutional neural networks considered in this paper allow for the bias term in each layer.

\begin{remark}
		 Upper bounds for the covering numbers of a single fully-connected layer or a single convolutional layer are also given in Lemma \ref{cn-fnn} and \ref{cn-cnn} in Appendix \ref{appendixa}.  In particular, let $X\in\mathbb{R}^{d_{in}}$ be an input vector of the layer with a bounded norm, i.e. $\Vert X\Vert_F=\Vert X\Vert_2<\infty$,  and let $\sigma$ be a $\kappa$-Lipschitz activation function.  For a convolutional layer with convolutional weights $C=(c_1,\ldots,c_r)\in\mathbb{R}^{r\times s}$ (with $r$ filters of size $s$,  each of which performs $m$ times operation, i.e. $d_{out}=r\times m$), induced weight matrix $W^c\in\mathbb{R}^{d_{out}\times d_{in}}$, and bias vector $b^c\in\mathbb{R}^{d_{out}}$ satisfying $\Vert C\Vert_F+\Vert b^c\Vert_F\leq a$, then the covering number of  the function class $\mathcal{F}^C_{a}(X):=\{\sigma(W^cX+b^c)\in\mathbb{R}^{d_{out}}:\Vert C\Vert_F+\Vert b^c\Vert_F\leq a\}$  under metric $\Vert\cdot\Vert_\infty$ with radius $\epsilon$ satisfies
		 $$\log \mathcal{N}(\mathcal{F}^C_{a}(X),\epsilon,\Vert\cdot\Vert_\infty)\leq r\times(s+m)\log\left\{1+\frac{2a \kappa(\sqrt{m}\Vert X\Vert_F+1)}{\epsilon}\right\}.$$
	To make a fair comparison, for a single fully-connected layer with weight matrix $W\in\mathbb{R}^{d_{out}\times d_{in}}$ and bias vector $b\in\mathbb{R}^{d_{out}}$, we consider the covering number of its output space subject to $\Vert W\Vert_F/\sqrt{m}+\Vert b\Vert_F\leq a$, i.e. $\mathcal{F}^F_{a}(X):=\{\sigma(WX+b)\in\mathbb{R}^{d_{out}}:\Vert W\Vert_F/\sqrt{m}+\Vert b\Vert_F\leq a\}$. It has been shown that  the upper bound of the covering number of $\mathcal{F}^F_a(X)$  under metric $\Vert\cdot\Vert_\infty$ with radius $\epsilon$ is
		 $$\log \mathcal{N}(\mathcal{F}^F_{a}(X),\epsilon,\Vert\cdot\Vert_\infty)\leq d_{out}\times (d_{in}+1)\log\left\{1+\frac{2a\kappa(\sqrt{m}\Vert X\Vert_F+1)}{\epsilon}\right\}.$$
		 It can be seen that the upper bound of the covering number of a single layer depends on the number of nonzero parameters in the layer. For a convolutional layer (CL) and a fully connected layer (FCL) with equal  input and output dimensions, as long as the number of filters $r\leq d_{out}$ and each filter size $s\leq d_{in}$, the covering number of CL will be smaller than that of FCL. In various image classification tasks (the data usually have special structures), FCLs generally performs no better than CLs in terms of their approximation powers or feature extraction ability, although
$d_{out} \gg r$ and $d_{in}\gg s.$ In such cases, the use of CNNs could dramatically reduce the networks complexities without sacrificing the performance and
boosts the computational efficiency.
\end{remark}

	Based on Theorem \ref{bound-cvnum-cnn}, we are now in a position to give an upper bound on the estimation error in terms of the covering number or the parameters of the function class $\mathcal{F}_{CNN}$.  We assume that the loss function $\phi$ satisfies the following assumption.

	\begin{assumption}\label{assumption1}
	The loss function $\phi:\mathbb{R}\to[0,\infty)$ is convex and classification-calibrated. Moreover, $\phi$ is $B_\phi$-Lipschitz on any bounded interval $M\subset[-\mathcal{B},\mathcal{B}]$, i.e.,
	$$\vert\phi(x)-\phi(y)\vert\leq B_\phi\vert x-y\vert,$$
	for any $x,y\in M\subset[-\mathcal{B},\mathcal{B}]$.
	\end{assumption}

The Lipschitz constants $B_\phi$ for several common $\phi$ loss functions are given in Table \ref{tab:2}.
	
	\begin{theorem}\label{bound-est}
Suppose that Assumption \ref{assumption1} holds.
		Let $\mathcal{F}_{CNN}$ be the function class defined  in (\ref{CNNs}),  $\mathcal{B}:=\sup_{f\in\mathcal{F}_{CNN}}\Vert f\Vert_\infty$ denote the bound of functions in $\mathcal{F}_{CNN}$ and let $\mathcal{S}=\sum_{l\in\mathcal{L}_F}(d_{l-1}+1)\times d_{l}+\sum_{l\in\mathcal{L}_C}(s_i+m_i)\times r_{l}$ denote the number of parameters for the networks in $\mathcal{F}_{CNN}.$
Then, the empirical $\phi$-risk minimizer $\hat{f}^\phi_n$ defined in (\ref{erm-cnn}) satisfies, for any $\delta \in (0, 1)$, with probability at least $1-\delta$,
			\begin{equation*}
				\sup_{f\in\mathcal{F}_{CNN}}\vert R^\phi(f)-R^\phi_n(f)\vert\leq \frac{8\sqrt{2}B_\phi\mathcal{B}\mathcal{S}^{1/2}L^{1/4}}{\sqrt{n}}+B_\phi\mathcal{B}\sqrt{\frac{2\log(1/\delta)}{n}},
			\end{equation*}
			and
			\begin{equation*}
				R^\phi(\hat{f}^\phi_n)-R^\phi(f^\phi_0)\leq \frac{16\sqrt{2}B_\phi\mathcal{B}\mathcal{S}^{1/2}L^{1/4}}{\sqrt{n}}+2B_\phi\mathcal{B}\sqrt{\frac{2\log(1/\delta)}{n}}+\inf_{f\in\mathcal{F}_{CNN}}R^\phi(f) -R^\phi(f^\phi_0).
			\end{equation*}
\end{theorem}
	
According to Lemma \ref{lemma2}, the estimation error $R^\phi(\hat{f}_n^\phi)-\inf_{f\in\mathcal{F}_{CNN}}R^\phi(f)$ is bounded by $2\sup_{f\in\mathcal{F}}\vert R^\phi(f)-R^\phi_n(f)\vert$ and  can be further bounded  by the complexity of $\mathcal{F}_{CNN}$ with high probability. In Theorem \ref{bound-est}, the complexity of $\mathcal{F}_{CNN}$ is measured by Rademacher complexity; by Dudley's integral,  it can be further contained by the covering number, which is represented in terms of  the network parameters $\mathcal{B},\mathcal{S}$ and $L$. This leads to an upper bound for the prediction/generalization error by
the sum of the estimation error and the approximation error of $\mathcal{F}_{CNN}$ to $f^\phi_0$.
	
\subsection{Approximation error}
	The approximation error $\inf_{f\in\mathcal{F}_{CNN}}R^\phi(f) -R^\phi(f^\phi_0)$  describes how well $R^\phi(f^\phi_0)$ can be approximated by CNNs in $\mathcal{F}_{CNN}$.
	The existing works on the approximation power of  neural networks mainly focus on
multi-layer perceptions, with a particular interest on the quantity $\inf_{f\in\mathcal{F}_{MLP}}\Vert f-f^\phi_0\Vert$ for a certain norm $\Vert\cdot\Vert$ and some smooth target function $f^\phi_0$ defined on a bounded compact set, e.g. $[0,1]^d$. In classification tasks, especially image classification, the support $\mathcal{X}$ of the input $X$ is exactly $[0,1]^d$. Thereafter,  we focus our discussion on those target functions $f^\phi_0$ defined on $\mathcal{X}=[0,1]^d.$
	
We derive an upper bound of the approximation error $\inf_{f\in\mathcal{F}_{CNN}}R^\phi(f) -R^\phi(f^\phi_0)$ by leveraging the recent powerful approximation theories for deep neural networks.
First, we
relate  $\inf_{f\in\mathcal{F}_{CNN}}R^\phi(f) -R^\phi(f^\phi_0)$ to $\inf_{f\in\mathcal{F}_{CNN}}\Vert f-f^\phi_0\Vert$.
Second, the target function $f^\phi_0$
may be  non-smooth or  unbounded,
 which poses extra difficulty in studying  $\inf_{f\in\mathcal{F}_{CNN}}\Vert f-f^\phi_0\Vert$,  	
	  as most of the existing studies on 
the approximation properties of deep neural networks assume smooth and bounded target functions.
 As shown in Table \ref{tab:1}, the target function $f^\phi_0$ for SVM is 
 non-continuous;  for the exponential and logistic loss,  their corresponding $f^\phi_0$ can be unbounded if
 $\eta(x)$ can be arbitrarily close to 0 or 1 for some $x\in\mathcal{X}.$
If the unboundedness problem cannot be solved satisfactorily,  then even under deterministic scenario,
 we have
$f_0^\phi(x)\in\{+\infty,-\infty\}$ for all $x\in\mathcal{X}$, which is nonsensical.
	
We overcome these difficulties by imposing mild and reasonable
 assumptions on the conditional probability $\eta$ and the loss function $\phi$.
	 \begin{assumption}\label{assumption2}
	 	\begin{itemize}
	 	\item [(a)]
The conditional probability $\eta(x)=\mathbb{P}\{Y=1\mid X=x\}$ is continuous on the support of $\mathcal{X}=[0,1]^d$ and the probability measure of $X$ is absolutely continuous with respect to the Lebesgue measure.
	  	\item [(b)] The loss function $\phi:\mathbb{R}\to[0,\infty)$ is continuously differentiable
on its support.
	 	\end{itemize}
	 \end{assumption}
{\color{black}
 	Under Assumptions \ref{assumption1}-\ref{assumption2}, we consider truncating the target function $f_0^\phi$ by a constant $T>0$, where $T$ may depend on $n$.
 Let $f_{0,T}^\phi$ be the truncated target function $f_0^\phi$, i.e.,
 \begin{equation}
 \label{fBphi}
  f_{0,T}^\phi(x)=\left\{\begin{array}{lr} f_0^\phi(x), \  \text{ if } \vert f_0^\phi(x)\vert\leq T,\\
   T, \ \ \ \ \ \ \, \text{ if } \vert f_0^\phi(x)\vert>T.
   \end{array}\right.
  \end{equation}
  Let $\bar{\mathbb{R}}=\mathbb{R}\cup\{+\infty,-\infty\}$ be the extended real line.  Then,  the approximation error can be decomposed into two terms that are easier to deal with.
 \begin{theorem}\label{bound-app}
	 	Suppose that Assumptions \ref{assumption1}-\ref{assumption2} hold and $T\leq\mathcal{B}$. Then,  the $\phi$-approximation error $\inf_{f\in\mathcal{F}_{CNN}}R^\phi(f) -R^\phi(f^\phi_0)$ with respect to the  loss function $\phi$ satisfies
	 	\begin{equation*}
	 		\inf_{f\in\mathcal{F}_{CNN}}R^\phi(f) -R^\phi(f^\phi_0)\leq B_\phi \inf_{f\in\mathcal{F}_{CNN}}\mathbb{E}\vert f(X)-f^\phi_{0,T}(X)\vert+\inf_{\vert a\vert\leq T}\phi(a)-\inf_{a\in\bar{\mathbb{R}}}\phi(a),
	 	\end{equation*}
 	where $f_{0,T}^\phi$ is the truncated  target function and  is continuous on $[0,1]^d$.
	 \end{theorem}
	
	By Theorem \ref{bound-app}, the $\phi$-approximation error $\inf_{f\in\mathcal{F}_{CNN}}R^\phi(f) -R^\phi(f^\phi_0)$ is bounded by the sum of two terms: $B_\phi\inf_{f\in\mathcal{F}_{CNN}}\mathbb{E}\vert f(X)-f^\phi_{0,T}(X)\vert$ and $\Delta_\phi(T):=\inf_{\vert a\vert\leq T}\phi(a)-\inf_{a\in\bar{\mathbb{R}}}\phi(a)$. We
 list the Lipschitz constant $B_\phi$ and $\Delta_\phi(T)$ for several commonly-used $\phi$ loss functions  in Table \ref{tab:2}.	
	\begin{table}[H]
		\caption{$\Delta_\phi(T)$, Lipschitz constant $B_\phi$ under different loss functions $\phi$ restricted to $[-\mathcal{B},\mathcal{B}]$ for $0<1\le T\le\mathcal{B}$, and  modulus of continuity $\omega_{f^\phi_{0,T}}$ of the truncated $f^\phi_{0}$.}
		\label{tab:2}
		\centering
		\resizebox{\textwidth}{!}{%
			\begin{tabular}{@{}lllll@{}}
				\toprule
				& Least squares     & SVM        &Exponential   		&Logistic          \\ \midrule
				$\phi(a)$           & $(1-a)^2$  & $\max\{1-a,0\}$      & $\exp(-a)$                                   & $\log\{1+\exp(-a)\}$                                 \\
				$f^\phi_0(x)$        & $2\eta-1$ & ${\rm sign}(2\eta-1)$ & $\frac{1}{2}\log(\frac{\eta}{1-\eta})$ & $\log(\frac{\eta}{1-\eta})$                    \\
				$B_\phi$ & $2\mathcal{B}$    & 1   & $\exp(\mathcal{B})$   & $1/\{\exp(-\mathcal{B})+1\}$ \\
				$\Delta_\phi(T)$		& 0 & 0 & $\exp(-T)$ & $\log\{1+\exp(-T)\}$\\ 	
				$\omega_{f^\phi_{0,T}}$ & $2\omega_{\eta}$ & *** & $\{\exp(T)+\exp(-T)\}^2\frac{\omega_{\eta}}{2}$ & $\{\exp(T/2)+\exp(-T/2)\}^2{\omega_{\eta}}$ \\
				 \bottomrule
			\end{tabular}
		}
		{\begin{flushleft}
		Note: $\eta(x)$ is written as $\eta$ for notational simplicity. "***" stands for no explicit form or not applicable.
		\end{flushleft}  }
	\end{table}
	
	It is worthwhile to note that the loss function $\phi$ of SVM does not satisfy Assumption \ref{assumption2} part (b);
	but under Assumption \ref{assumption2} part (a), we can obtain a  minimizer $f_0^\phi$
that is a step function. To be specific,
for the SVM risk function, $f_0^\phi(x)$ can be defined as any value in $[-1,1]$ when $\eta(x)=1/2$. Hence,  we define $f_0^\phi(x)={\rm sign}(2\eta(x)-1)$ if $\eta(x)\not=1/2$, otherwise
define $f_0^\phi(x)=c$ for some $c\in[-1,1]$ such that $f_0^\phi$ is a step function on $[0,1]^d$.
	
	By Theorem \ref{bound-app}, the approximation error can be controlled by  properly choosing  $T$ to bound $\Delta_\phi(T)$ and $B_\phi\inf_{f\in\mathcal{F}_{CNN}}\mathbb{E}\vert f(X)-f^\phi_{0,T}(X)\vert$. For the latter, we can apply existing approximation results of neural networks to handle it. Based on the existing approximation theories, the upper bound of $\inf_{f\in\mathcal{F}_{CNN}}\mathbb{E}\vert f(X)-f^\phi_{0,T}(X)\vert$ depends on $\mathcal{F}_{CNN}$ through its parameters and related to $f^\phi_0$ through its modulus of continuity. The modulus of continuity $\omega_f$ of a function $f:[0,1]^d\to\mathbb{R}$ is defined by
\begin{equation}
\label{omega}
\omega_f(r) :=\sup\{\vert f(x)-f(y)\vert:x,y\in[0,1]^d,\Vert x-y\Vert_2\leq r\}, {\rm for\ any\ } r\geq0.
\end{equation}
For a 
uniformly continuous function $f$, $\lim_{r\to0}\omega_f(r)=\omega_f(0)=0$.  In addition, based on
the modulus of continuity,
different equi-continuous families of functions can be defined. For instance, the modulus $\omega_f(r)= Lr$ describes the $L$-Lipschitz continuity; the modulus $\omega_f(r)=\lambda r^\alpha$ with $\lambda,\alpha>0$ describes the H\"older continuity and so on.

In our classification problem, we can view $f_0^\phi=g_0^\phi\circ\eta$ as a composition of a function $g_0^\phi$ and the conditional probability function $\eta$ defined in (\ref{eta}).
	For example,  $g_0^\phi(\eta)=2\eta-1$ for the least squares loss;  $g_0^\phi(\eta)=\log\{\eta/(1-\eta)\}$ for the logistic loss;   $g_0^\phi(\eta)=\log(\eta/(1-\eta))/2$ for the exponential loss;  and $g_0^\phi(\eta)={\rm sign}(2\eta-1)$ for the SVM. For the composite function $f_{0}^\phi=g_0^\phi\circ\eta$,
its modulus of continuity is given by $\omega_{f_0^\phi}=\omega_{g_0^\phi}\circ\omega_{\eta}$. Moreover, the modulus of continuity of the truncated function $f^\phi_{0,T}$ is bounded by that of the target function $f^\phi_{0}$. The modulus of continuity of $f^\phi_{0,T}$ for the least squares, the logistic,  the exponential, and the SVM  are given in Table \ref{tab:2}, which are
 a function of $\omega_{\eta}$, the modulus of continuity of $\eta$.
	Assumptions \ref{assumption1}-\ref{assumption2} imply  that $f^\phi_{0,T}$ is uniformly continuous, the proof is included in the proof of Theorem \ref{bound-app} in Appendix).
Therefore, our results are  expressed
in terms of the modulus of continuity of $f^\phi_{0,T}$.
}

\section{Non-asymptotic error bounds}\label{sec4}
	Theorems \ref{bound-est} and \ref{bound-app} provide the theoretical basis for establishing
	the consistency and non-asymptotic excess risk bounds.
	In this section, we first give an explicit upper bounds for the excess $\phi$-risk $R^\phi(\hat{f}_n)-R^\phi(f^\phi_0)$ with a general loss function $\phi$ satisfying Assumptions \ref{assumption1} and \ref{assumption2}. We also show that classification with CNN is able to circumvent the curse of dimensionality under an approximate lower-dimensional manifold support assumption on the distribution of the input data $X$.

\subsection{Non-asymptotic excess risk bound}
{\color{black}
\begin{theorem}[Non-asymptotic excess $\phi$-risk bound]\label{gerrorbound}
		Suppose that Assumptions \ref{assumption1} and \ref{assumption2} hold. For any $M, N\in\mathbb{N}^+$, let $\mathcal{F}_{CNN}$ be the class of CNNs defined in (\ref{CNNs}) with $B_a\leq\max\{M^2N^2,2^{M+1}\}$, $ f^\phi_{0,T}(\textbf{0})+\omega_{f^\phi_{0,T}}(\sqrt{d})\le\mathcal{B}$, depth $L\leq12M+14$ and width $\mathcal{W}\leq\max\{4d\lfloor N^{\frac{1}{d}}\rfloor+3d,12N+8\}$,
		where $\lfloor a \rfloor$ denotes the largest integer strictly smaller than $a$. Then,  for any $\delta\in (0, 1)$, with probability at least $1-\delta$, the ERM $\hat{f}^\phi_n$ defined in (\ref{erm-cnn}) satisfies
		\begin{align}
\label{gerrorbound-inq}
			R^\phi(\hat{f}^\phi_n)-R^\phi(f_0^\phi)\leq&
\text{EstError} + \text{AppError},
		\end{align}
where the \textit{estimation error}
\[
\text{EstError} =
16\sqrt{2}B_\phi\mathcal{B}\frac{\mathcal{S}^{1/2}L^{1/4}}{\sqrt{n}}+
2B_\phi\mathcal{B}\frac{\sqrt{2\log(1/\delta)}}{\sqrt{n}},
\]
and the \textit{approximation error}
\[
\text{AppError}= 18\sqrt{d}B_\phi \omega_{f^\phi_{0,T}}(N^{-2/d}M^{-2/d})+\inf_{\vert a\vert\leq T}\phi(a)-\inf_{a\in\bar{\mathbb{R}}}\phi(a).
\]
\end{theorem}
}

The upper bound of the excess $\phi$-risk $R^\phi(\hat{f}^\phi_n)-R^\phi(f_0^\phi)$ in Theorem \ref{gerrorbound} is a sum of two error terms, the estimation error and the approximation error.
It is worthwhile to note the following two aspects.

\begin{itemize}
\item First, the error bound (\ref{gerrorbound-inq})
is  non-asymptotic and explicit  in the sense that  no unclearly defined 
constant are involved. The prefactor $18\sqrt{d}B_\phi$ in the upper bound of approximation error (the first item in \textit{AppError}) depends on the dimension $d$ sub-linearly, different from the exponential dependence in many existing neural network approximation results.
\item
Second, the approximation rate $(NM)^{-2/d}$ is in terms of the width $\mathcal{W}\leq\max\{4d\lfloor N\rfloor^{1/d}+3d,12N+8\}$ and depth $L\leq12M+14$, rather than just the size $\mathcal{S}$ of the network. This offers some insights into how the excess risk relates to the network structure.
\end{itemize}

To achieve the optimal error rate, we need to balance the trade-off between the estimation error and the approximation error. On one hand, the bound for the estimation error \text{EstError} 
increases with the size and the depth 
of $\mathcal{F}_{CNN}.$
On the other hand, the  bound for the approximation error \textit{AppError}
decreases with the depth and the width of  $\mathcal{F}_{CNN}.$
Notably, the choice of the bound $\mathcal{B}$ is also crucial to balance the errors and achieve good rate of convergence. As a result, proper design of the network architecture is of key importance to ensure good convergence rate of the excess $\phi$-risk.
	
Theorem \ref{gerrorbound}	immediately yields the following consistency result for the ERM 	$\hat{f}^\phi_n.$

{\color{black}
\begin{theorem}[Consistency]\label{consistency}
		Suppose that Assumptions \ref{assumption1} and \ref{assumption2} hold. Let $\mathcal{F}_{CNN}$ be
		the function class of CNNs defined in (\ref{CNNs}) with
$f^\phi_{0,T}(\textbf{0})+\omega_{f^\phi_{0,T}}(\sqrt{d})\le\mathcal{B}.$
If
		
		\begin{align*}
		\mathcal{S}\to\infty, \  \sqrt{d}B_\phi\omega_{f^\phi_{0,T}}(\mathcal{S}^{-2/d})\to0, \  \frac{B^\phi\mathcal{B}\mathcal{S}^{1/2}L^{1/4}}{\sqrt{n}}\to0 {\rm\ and\ } \inf_{\vert a\vert\leq T}\phi(a)-\inf_{a\in\bar{\mathbb{R}}}\phi(a)\to0,
		\end{align*}
		as $n\to\infty$, then the ERM $\hat{f}^\phi_n$ w.r.t the loss $\phi$ defined in (\ref{erm-cnn}) and the 0-1 loss excess risk of $\hat{h}_n={\rm sign}(\hat{f}^\phi_n)$ satisfy
		\begin{align*}
			R^\phi(\hat{f}^\phi_n)-R^\phi(f_0^\phi)\to0 {\rm \quad and\quad } R(\hat{h}_n)-R_0\to0
		\end{align*}
		in probability.
\end{theorem}
}
	
\subsection{Circumventing the curse of dimensionality}\label{sec4.2}
		
In many modern machine learning problems,
the ambient dimension $d$ of the input data could be very large,  which results in an extremely slow convergence rate even if there is a large sample size.  This fact is known as the curse of dimensionality.
%
However, many types of data have a low-dimensional latent structure, that is, although the ambient dimension $d$  is large, the distribution of the data is approximately support on a low-dimensional subset of $\mathbb{R}^d$ . Apparently, it is desirable to incorporate such a latent structure in the theoretical analysis.
In Theorem \ref{bound-app}, the approximation error $\inf_{f\in\mathcal{F}_{CNN}}R^\phi(f) -R^\phi(f^\phi_0)$ is actually defined with respect to the probability measure $\mathbb{P}$.
In light of this,  a proper and realistic assumption on the support of $X$ can help lessen the curse of dimensionality. Although $f^\phi_0$ is fixed and its domain is high dimensional, when the support of $X$ is concentrated on a neighborhood of a low-dimensional manifold, the upper bound of the approximation error can be substantially improved
in terms of the exponent of the convergence rate  \citep{shen2019deep}.

	\begin{assumption}
		\label{assumption3}
		The covariate $X$ is supported on $\mathcal{M}_\rho$, a $\rho$-neighborhood of $\mathcal{M}\subset[0,1]^d$, where $\mathcal{M}$ is a compact $d_\mathcal{M}$-dimensional Riemannian submanifold and $\mathcal{M}_\rho=\{x\in[0,1]^d: \inf\{\Vert x-y\Vert_2: y\in\mathcal{M}\}\leq \rho\}$ for $\rho\in(0,1)$.
	\end{assumption}

In real-world applications, data are hardly observed to locate on an \textit{exact manifold}, instead they could be more realistically viewed as
consisting of a latent part supported on a low-dimensional manifold $\mathcal{M}$ plus noises.  Therefore, Assumption \ref{assumption3} is more reasonable compared with the exact manifold assumption assumed in  \cite{schmidt2019deep,nakada2019adaptive,chen2019nonparametric}.

Before stating the main theorem in this subsection, we first present a result on the approximation power of
CNNs under an approximate low-dimensional manifold support assumption. This result is of independent interest and can be useful in other problems that involve the use of CNNs.
Let $f^\phi_{0,T}$ be defined as in (\ref{fBphi}).

\begin{lemma}
\label{approx-lowDsupport}
		Suppose that Assumptions \ref{assumption1}-\ref{assumption3} hold.
{\color{black} For any $M, N\in\mathbb{N}^+$, let $\mathcal{F}_{CNN}$ be
		the class of CNNs defined in (\ref{CNNs}) with $B_a\leq\max\{M^2N^2,2^{M+1}\}$, $f^\phi_{0,T}(\textbf{0})+\omega_{f^\phi_{0,T}}(\sqrt{d})\le\mathcal{B}$, depth $L\leq12M+14$} and
{\color{black} width $\mathcal{W}\leq\max\{4d_\varepsilon\lfloor N^{\frac{1}{d_\varepsilon}}\rfloor+3d_\varepsilon,12N+8\}$, where $d_\varepsilon=O(d_\mathcal{M}\frac{\log(d/\varepsilon)}{\varepsilon^2})$ for $\varepsilon \in (0, 1).$}
Suppose the radius of the neighborhood $\rho$ in Assumption \ref{assumption3} satisfies $\rho \leq N^{-2/d_\varepsilon}M^{-2/d_\varepsilon}(1-\varepsilon)/\{2(\sqrt{d/d_\varepsilon}+1-\varepsilon)\}.$
Then, we have
	\begin{equation*}
		\inf_{f \in \mathcal{F}_{\text{CNN}}} \mathbb{E}\vert f(X) -f^\phi_{0,T}(X)\vert \leq (18\sqrt{d_\varepsilon}+2)\omega_{f^\phi_{0,T}}(N^{-2/d_\varepsilon}M^{-2/d_\varepsilon}).
	\end{equation*}
\end{lemma}

The following theorem provides a non-asymptotic excess risk bound under the approximate low-dimensional
manifold  assumption.
	
\begin{theorem}[Circumventing the curse of dimensionality]\label{gerrorbound-lowdim}
		Suppose that Assumptions \ref{assumption1}-\ref{assumption3} hold.
{\color{black} For any $M, N\in\mathbb{N}^+$, let $\mathcal{F}_{CNN}$ be
		the class of CNNs defined in (\ref{CNNs}) with $B_a\leq\max\{M^2N^2,2^{M+1}\}$, $f^\phi_{0,T}(\textbf{0})+\omega_{f^\phi_{0,T}}(\sqrt{d})\le\mathcal{B}$, depth $L\leq12M+14$} and
{\color{black} width $\mathcal{W}\leq\max\{4d_\varepsilon\lfloor N^{\frac{1}{d_\varepsilon}}\rfloor+3d_\varepsilon,12N+8\}$, where $d_\varepsilon=O(d_\mathcal{M}\frac{\log(d/\varepsilon)}{\varepsilon^2})$ for $\varepsilon \in (0, 1).$}
Suppose the radius of the neighborhood $\rho$ in Assumption \ref{assumption3} satisfies $\rho \leq N^{-2/d_\varepsilon}M^{-2/d_\varepsilon}(1-\varepsilon)/\{2(\sqrt{d/d_\varepsilon}+1-\varepsilon)\}.$
Then for any $\delta \in (0, 1)$, with probability at least $1-\delta$, the ERM $\hat{f}^\phi_n$ defined in (\ref{erm-cnn}) satisfies
%
%
		\begin{align}
\label{gerrorbound-lowdim-inq}
			R^\phi(\hat{f}^\phi_n)-R^\phi(f_0^\phi)\leq&
\text{EstError}_* + \text{AppError}_*,
		\end{align}
where the \textit{estimation error}
\[
\text{EstError}_* =16\sqrt{2}B_\phi\mathcal{B}\frac{\mathcal{S}^{1/2}L^{1/4}}{\sqrt{n}}+
2B_\phi\mathcal{B}\frac{\sqrt{2\log(1/\delta)}}{\sqrt{n}},
\]
and the \textit{approximation error}
\[
\text{AppError}_*=(18\sqrt{d_\varepsilon}+2)B_\phi \omega_{f^\phi_{0,T}}(N^{-2/d_\varepsilon}M^{-2/d_\varepsilon})+\inf_{\vert a\vert\leq T}\phi(a)-\inf_{a\in\bar{\mathbb{R}}}\phi(a).
\]
	\end{theorem}
For a high-dimensional input $X$ with a large $d$,  $d_\varepsilon=O(d_\mathcal{M}{\log(d/\varepsilon)}/{\varepsilon^2})$ satisfies
$d_\mathcal{M}$$\leq d_\varepsilon<d$ for $\varepsilon\in(0,1)$.
Comparing with the bound (\ref{gerrorbound-inq}) in Theorem \ref{gerrorbound}, we see that
$\textit{EstError}_*=\textit{EstError}$, that is, the estimation error does not change under the approximate
low-dimensional manifold assumption. The change is in the approximation error term $\textit{AppError}_*$.
Comparing with $\textit{AppError}$ in (\ref{gerrorbound-inq}) of Theorem \ref{gerrorbound}, we see that it only depends on $d_{\varepsilon},$  which leads to a much faster convergence rate.
Therefore,
Theorem \ref{gerrorbound-lowdim} significantly improves the error bound in comparison with the error bound in Theorem \ref{gerrorbound} without any assumption on the support of $X$.
Theorem \ref{gerrorbound-lowdim} shows that classification with CNNs can circumvent
the curse of dimensionality if the input data $X$ is supported on an approximate low-dimensional manifold.
	
\section{Examples}
\label{examples}
In this section, we  illustrate the applications of Theorems \ref{gerrorbound} and  \ref{gerrorbound-lowdim}
 to obtaining convergence rates of excess risk in classification.
We apply the general excess risk bounds established in these theorems to several important classification methods with CNNs in details when the specific form of $\phi$ is given,  including those with the least squares, the logistic, the exponential, and the SVM losses.

We make
mild assumptions on the modulus of continuity of
the condition probability function
$\eta$ defined in (\ref{eta}), i.e.,  there exist $\lambda \ge 0$ and $\alpha>0$ such that $\vert \eta(x)-\eta(y)\vert\leq \lambda\Vert x-y\Vert_2^\alpha$ for any $x,y\in\mathcal{X}\subset[0,1]^d$. With  the H\"older-continuity assumption on $\eta$, its modulus of continuity $\omega_{\eta}(r)=\lambda r^\alpha$
based on its definition (\ref{omega}).
Typically,  as in many machine leaning tasks, we  assume the input dimension $d\gg \alpha$ by default.
For the examples given below, the network is chosen to be fixed-width networks, thus the optimal convergence rate can be achieved with the minimal size (number of parameters) due to the fact that the network size must satisfy $\mathcal{S}\leq\mathcal{W}^2 L$. A detailed discussion on this point can be found in \citet{jiao2021dnr}.
	
\subsection{Example 1: Least squares}
	For the least squares loss $\phi(a)=(1-a)^2$,  the approximation error $$\inf_{f\in\mathcal{F}_{CNN}}R^\phi(f) -R^\phi(f^\phi_0)= \inf_{f\in\mathcal{F}_{CNN}}\mathbb{E}\vert f(X)-f^\phi_{0}(X)\vert^2.$$
{\color{black}
	Let $\mathcal{F}_{CNN}$ be the class of CNNs  defined in (\ref{CNNs}) with $1+2\lambda d^{\alpha/2}\le\mathcal{B}$,
	with the  number of layers $L\leq12\lfloor n^{d/2(d+2\alpha)}\rfloor+14$, width $\mathcal{W}\leq\max\{7d,20\}$ and size (number of parameters) $\mathcal{S}\leq\mathcal{W}^2 L\leq \max\{49d^2,400\}\times\{\lfloor n^{d/2(d+2\alpha)}\rfloor+14\}.$
	Suppose that Assumptions \ref{assumption1} and \ref{assumption2} hold.
	Theorem \ref{gerrorbound} implies that,   with probability at least $1-\exp\{-n^{(d-2\alpha)/(d+2\alpha)}\}$, the excess $\phi$-risk of the ERM $\hat{f}^\phi_n$ defined in (\ref{erm-cnn}) satisfies
	\begin{align}
		\label{lsrisk}
		R^\phi(\hat{f}^\phi_n)-R^\phi(f_0^\phi)\leq& \Big\{(32\sqrt{2}c+4\sqrt{2})\mathcal{B}^2+4\cdot18^2\lambda^2d \mathcal{B}(\max\{7d,20\})^{-\frac{4\alpha}{d}}\Big\}n^{-\frac{2\alpha}{(d+2\alpha)}},
	\end{align}
	where $c=O(d)$ is a constant independent of $n,\mathcal{B},\alpha$ and $\lambda$.
	Alternatively,  (\ref{lsrisk}) can be simply written as $$R^\phi(\hat{f}^\phi_n)-R^\phi(f_0^\phi)\leq C(d)n^{-\frac{2\alpha}{(d+2\alpha)}}$$ for some constant $C(d)=O(d(4\lambda^2d^\alpha+1))$ independent of $n,\mathcal{B}$. Here,  the convergence rate achieves the optimal minimax rate in \cite{stone1982optimal}; more importantly, the prefactor $C(d)$ depends on the dimension $d$ at a rate no more than $O(d^{1+\alpha})$,  which is much improved than the exponential dependence $O(2^d)$ in
the existing works.

	In addition, suppose the approximate low-dimensional manifold Assumption \ref{assumption3} also holds and for any $ \varepsilon \in (0,1),$ the radius of the neighborhood $\rho$ in Assumption \ref{assumption3} satisfies $\rho \leq N^{-2/d_\varepsilon}M^{-2/d_\varepsilon}(1-\varepsilon)/\{2(\sqrt{d/d_\varepsilon}+1-\varepsilon)\},$
	where $d_\varepsilon=O(d_\mathcal{M}{\log(d/\varepsilon)}/{\varepsilon^2})$ is an integer satisfying $d_\mathcal{M}$$\leq d_\varepsilon \ll d.$
	Then by Theorem \ref{gerrorbound-lowdim}, the rate of convergence can be improved to
	$$R^\phi(\hat{f}^\phi_n)-R^\phi(f_0^\phi)\leq C(d_\varepsilon)n^{-\frac{2\alpha}{(d_{\varepsilon}+2\alpha)}},$$
	for some constant $C(d_\varepsilon)=O(d_\varepsilon(4\lambda^2d_\varepsilon^\alpha+1))$ independent of $n$ and $\mathcal{B}$. Again, here the prefactor depends on $d$ at most polynomially.
}

\subsection{Example 2:
Logistic loss}
For the logistic loss function $\phi(a)=\log\{1+\exp(-a)\}$, the  bound of  excess risk is not tight enough, as the minimizer of $\phi$-risk
		$\log\{{\eta(x)}/{1-\eta(x)}\}$ is unbounded with the modulus of continuity of truncated $f^\phi_{0,T}$ being $\{\exp(T/2)+\exp(-T/2)\}^2{\omega_{\eta}}$ and the error due to the truncation $\Delta_\phi(T)=\log\{1+\exp(-T)\}$. This requires a suitable choice of $T$ to control the two items simultaneously. A feasible choice is $T=O(\log\log(n))$ where $n$ is the sample size.

Let $\mathcal{F}_{CNN}$ be the class of CNNs defined in (\ref{CNNs}).
With a suitable number of layers, width and size, it holds that with high probability, the excess $\phi$-risk of the ERM $\hat{f}^\phi_n$ defined in (\ref{erm-cnn}) satisfies $R^\phi(\hat{f}_n)-R^\phi(f_0^\phi)\leq C/\log(n)$ for some constant $C>0$  not depending on $n$.
		
A practical way to improve the  
rate of convergence is to use a modified logistic loss function, i.e. $\phi(a)=\max\{\log\{1+\exp(-a)\},\tau\}$ for some small $\tau>0$. With the modified logistic loss, the minimum of $\phi$ can be achieved at a finite number $a^*(\tau)=-\log\{\exp(\tau)-1\}$, and the
		minimizer  $f^\phi_0(x)$ will be a truncated version of $\log[\eta(x)/\{1-\eta(x)\}]$ (truncated by $a^*(\tau)=-\log\{\exp(\tau)-1\}$). In light of this,
			under the modified logistic loss with $\tau=\log\{1+\exp(-T)\}$, the corresponding measurable minimizer defined in (\ref{ermphi}) is $f^\phi_{0,T}$, the truncated version of the minimizer $f_0^\phi$ under the original logistic loss; see Table \ref{tab:2}. As a consequence, under the modified logistic loss, $\Delta_\phi(T)=0$ by definition since the infimum of the modified logistic loss can be achieved within $[-T,T]$. Note that

{\color{black} Suppose Assumptions \ref{assumption1} and \ref{assumption2} hold.
	Let $\mathcal{F}_{CNN}$ be the class of CNNs  defined in (\ref{CNNs}) with $\min(\vert\log[\eta(\textbf{0})/\{1-\eta(\textbf{0})\}]\vert,T\}+4\exp(T)\lambda d^{\alpha/2}\le\mathcal{B}$, the number of layers $L\leq12\lfloor n^{2d/(3d+8\alpha)}\rfloor+14$, width $\mathcal{W}\leq\max\{7d,20\}$ and size $\mathcal{S}\leq\mathcal{W}^2 L\leq \max\{49d^2,400 \}\times(\lfloor n^{2d/(3d+8\alpha)}\rfloor+14)$.
	
	Theorem \ref{gerrorbound} implies that,  with probability at least $1-\exp\{-n^{d/(d+8\alpha/3)}\}$, the excess $\phi$-risk of the ERM
	$\hat{f}^\phi_n$ defined in (\ref{erm-cnn}) satisfies
	\begin{align}
		\label{lgrisk}
		R^\phi(\hat{f}^\phi_n)-R^\phi(f_0^\phi)\leq& \Big\{(16\sqrt{2}c+2\sqrt{2})\mathcal{B}+72\lambda\exp(T)\sqrt{d} (\max\{7d,20\})^{-\frac{2\alpha}{d}}\Big\}n^{-\frac{\alpha}{0.75d+2\alpha}},
	\end{align}
	where $c$ is a constant independent of $n,\mathcal{B},\alpha,\lambda$, and $c$ is independent of $d$ if $d\leq2$ otherwise $c=O(d)$.
	Alternatively,  (\ref{lgrisk}) can be simply written as 		
	$$R^\phi(\hat{f}^\phi_n)-R^\phi(f_0^\phi)\leq C(T,d)n^{-\frac{\alpha}{0.75d+2\alpha}}, $$
	where $C(T,d)=O(d(1+4\lambda d^{\alpha/2})\exp(T))$ is a constant independent of $n,\mathcal{B}$. Here the prefactor $C(T,d)$ depends on the dimension $d$ at a rate no more than $O(d^{\alpha/2+1})$.

	In addition, suppose the approximate low-dimensional manifold Assumption \ref{assumption3} also holds and for any $ \varepsilon \in (0,1),$ the radius of the neighborhood $\rho$ in Assumption \ref{assumption3} satisfies
	$\rho \leq N^{-2/d_\varepsilon}
	M^{-2/d_\varepsilon}(1-\varepsilon)/\{2(\sqrt{d/d_\varepsilon}+1-\varepsilon)\},$
	where $d_\varepsilon=O(d_\mathcal{M}{\log(d/\varepsilon)}/{\varepsilon^2})$ is an integer satisfying $d_\mathcal{M}$$\leq d_\varepsilon \ll d.$
	Then Theorem \ref{gerrorbound-lowdim} shows that the rate of convergence can be improved to
	$$R^\phi(\hat{f}^\phi_n)-R^\phi(f_0^\phi)\leq C(T,d_\varepsilon)n^{-\frac{\alpha}{0.75d_{\varepsilon}+2\alpha}}, $$
	where $C(T,d_\varepsilon)=O(d_\varepsilon(1+4\lambda d_\varepsilon^{\alpha/2})\exp(T))$ is a constant independent of $n$ and $\mathcal{B}$.
}
	
\subsection{Example 3: 
Exponential loss}
		For the exponential loss function $\phi(a)=\exp(-a)$, the  upper bound of the excess risk	
			  suffers from  the same problem as the logistic loss, as the minimizer of the $\phi$-risk $\log\{{\eta(x)}/{1-\eta(x)}\}/2$ is unbounded.
			  Furthermore, for the exponential loss, the Lipschitz constant and the modulus of continuity grows exponentially fast as $T$ grows, which makes it harder to balance  $B_\phi$, $\omega_{f^\phi_{0,T}}$ and $\Delta_\phi(T)$. Similar to the logistic loss, we consider a modified exponential loss function $\phi(a)=\max\{\exp(-a),\tau\}$ for some small $\tau>0$, which
			   results in a bounded measurable minimizer $f^\phi_0(x)$.
			   Under modified exponential loss with $\tau=\exp(-T)$, its corresponding measurable minimizer defined in (\ref{ermphi}) is $f^\phi_{0,T}$, the truncated version of the minimizer $f^\phi_0$  under the original exponential loss; see Table \ref{tab:2}. Also, under modified exponential loss, $\Delta_\phi(T)=0$ by definition.

{\color{black} Suppose Assumptions \ref{assumption1} and \ref{assumption2} hold.  Let $\mathcal{F}_{CNN}$ be the class of CNNs  defined in (\ref{CNNs}) with $\min(\vert\log[\eta(\textbf{0})/\{1-\eta(\textbf{0})\}]\vert/2,T\}+2\exp(2T)\lambda d^{\alpha/2}\le\mathcal{B}$,
	the number of layers $L\leq12\lfloor n^{2d/(3d+8\alpha)}\rfloor+14$, width $\mathcal{W}\leq\max\{7d,20\}$ and size $\mathcal{S}\leq\mathcal{W}^2 L\leq \max\{49d^2,400\}\times(\lfloor n^{2d/(3d+8\alpha)}\rfloor+14).$
	Theorem \ref{gerrorbound} implies that, with probability at least $1-\exp\{-n^{d/(d+8\alpha/3)}\},$
	the excess $\phi$-risk of the ERM $\hat{f}^\phi_n$ defined in (\ref{erm-cnn}) satisfies
	\begin{align}
		\label{exprisk}
		R^\phi(\hat{f}^\phi_n)-R^\phi(f_0^\phi)\leq& \Big\{(16\sqrt{2}c+2\sqrt{2})\mathcal{B}\exp(\mathcal{B})+36\lambda\exp(2T+\mathcal{B})\sqrt{d} (\max\{7d,20\})^{-2\alpha/d}\Big\} \nonumber \\
		&\hskip 6cm	\times n^{-\alpha/(0.75d+2\alpha)},
	\end{align}
	where  $c$ is a constant independent of $n,\mathcal{B},\alpha$ and $\lambda$, and is independent of $d$ if $d\leq2$ otherwise $c=O(d)$. 		
	Alternatively, (\ref{exprisk}) can be recast as
	$$R^\phi(\hat{f}^\phi_n)-R^\phi(f_0^\phi)\leq C(T,d)n^{-\alpha/(0.75d+2\alpha)}, $$ where $C(T,d)=O(d(1+2\lambda d^{\alpha/2})\exp(2T)\exp\{3\exp(2T)(1+2\lambda d^{\alpha/2})\})$ is a constant independent of $n$ and $\mathcal{B}$. Here the prefactor $C(T,d)=O(d^{\alpha/2+1}\exp(d^{\alpha/2}))$ depends on the dimension $d$ exponentially.

	In addition, suppose the approximate low-dimensional manifold Assumption \ref{assumption3} also holds and for any $ \varepsilon \in (0,1),$ the radius of the neighborhood $\rho$ in Assumption \ref{assumption3} satisfies
	$\rho \leq N^{-2/d_\varepsilon}
	M^{-2/d_\varepsilon}(1-\varepsilon)/\{2(\sqrt{d/d_\varepsilon}+1-\varepsilon)\},$
	where $d_\varepsilon=O(d_\mathcal{M}{\log(d/\varepsilon)}/{\varepsilon^2})$ is an integer satisfying $d_\mathcal{M}$$\leq d_\varepsilon\ll d.$
	Then Theorem \ref{gerrorbound-lowdim} shows that the rate of convergence can be improved to
	$$R^\phi(\hat{f}^\phi_n)-R^\phi(f_0^\phi)\leq C(T,d_\varepsilon)n^{-\alpha/(0.75d_{\varepsilon}+2\alpha)},$$
	where $C(T,d_\varepsilon)=O(d_\varepsilon(1+2\lambda d_\varepsilon^{\alpha/2})\exp(2T)\exp\{3\exp(2T)(1+2\lambda d_\varepsilon^{\alpha/2})\})$ is a constant independent of $n$ and $\mathcal{B}$.
}
	
\subsection{Example 4: SVM}
		For the support vector machine (SVM), the loss function $\phi(a)=\max\{1-a,0\}$ is not differentiable at $a=1$ and thus the $\phi$-risk minimizer $f_0^\phi(x)={\rm sign}(2\eta(x)-1)$ may not be continuous even though $\eta$ is continuous, as $f_0^\phi(x)$ is discontinuous when $\eta(x)=1/2$.
		To tackle this problem, we additionally impose the low noise condition on $\eta$ (\cite{mammen1999smooth,tsybakov2004optimal}), i.e. there exist $c_{\text{noise}}>0$ and $q\in[0,\infty]$ such that for any $t>0$,
$$\mathbb{P}(\vert 2\eta(X)-1\vert\leq t)\leq c_{\text{noise}}t^q,$$
where the constant $q$ is called the noise exponent.

{\color{black} Suppose Assumptions \ref{assumption1} and \ref{assumption2} hold.
	Let $\mathcal{F}_{CNN}$ be the class of CNNs  defined in (\ref{CNNs}) with $\mathcal{B}\ge1$
	with  the number of layers $L\leq12\lfloor n^{2d/\{3d+8\alpha(q+1)\}}\rfloor+15$, width $\mathcal{W}\leq\max\{7d,20\}$ and size (number of parameters) $\mathcal{S}\leq\mathcal{W}^2 L\leq \max\{49d^2,400\}\times[\lfloor n^{2d/\{3d+8\alpha(q+1)\}}\rfloor+15]$.
	Theorem \ref{gerrorbound-lowdim} implies that,  with probability at least $1-\exp[-n^{3d/\{3d+8\alpha(q+1)\}}]$, the excess $\phi$-risk of the ERM  $\hat{f}^\phi_n$ defined in (\ref{erm-cnn}) satisfies
	\begin{align*}
		R^\phi(\hat{f}_n)-R^\phi(f_0^\phi)\leq& \Big\{16\sqrt{2}c+2\sqrt{2}+144\times4^q\lambda c_{noise}\sqrt{d} (\max\{7d,20\})^{-\frac{2\alpha}{d}}\Big\}n^{-\frac{4\alpha(q+1)}{3d+8\alpha(q+1)}},
	\end{align*}
	where $c$ is a constant independent of $n,\mathcal{B},\alpha$ and $\lambda$, and is independent of $d$ if $d\leq2$ otherwise $c=O(d)$, or simply
	$$R^\phi(\hat{f}_n)-R^\phi(f_0^\phi)\leq C(d)n^{-4\alpha(q+1)/\{3d+8\alpha(q+1)\}},$$
	where $C(d)$ is a constant independent of $n,\mathcal{B}$ and $\alpha$, and $C(d)=O(\sqrt{d})$ if $d\leq2$ otherwise $C=O(d)$.
}We note that the excess risk bound depends on $q$, the noise exponent. {\color{black} When $q=0$ (high noise), the convergence rate with respect to the sample size is $n^{-\alpha/(0.75d+2\alpha)}$, which is exactly the same as Modified Logistic or exponential examples; When $q=+\infty$ (no noise), the rate will be significantly improved to $n^{-1/2}$.}
A similar result can be found in Theorem 3.3 of \cite{kim2021fast}.

{\color{black}
In addition, suppose the approximate low-dimensional manifold Assumption \ref{assumption3} also holds and for any $ \varepsilon \in (0,1),$ the radius of the neighborhood $\rho$ in Assumption \ref{assumption3} satisfies
$\rho \leq N^{-2/d_\varepsilon}
M^{-2/d_\varepsilon}(1-\varepsilon)/\{2(\sqrt{d/d_\varepsilon}+1-\varepsilon)\},$
where $d_\varepsilon=O(d_\mathcal{M}{\log(d/\varepsilon)}/{\varepsilon^2})$ is an integer satisfying $d_\mathcal{M}$$\leq d_\varepsilon\ll d.$
Then Theorem \ref{gerrorbound-lowdim} shows the rate of convergence can be improved to
\[
R^\phi(\hat{f}_n)-R^\phi(f_0^\phi)\leq C \sqrt{d_\varepsilon}n^{-4\alpha(q+1)/\{3d_{\varepsilon}+8\alpha(q+1)\}}.
\]
}
		

The proofs for the above four examples are deferred to Appendix.
{\color{black}
It is interesting to note that the excess bounds are different for different loss functions. First, the exponents in the risk bounds depend on the dimensionality $d$ differently. For least squares, the approximation error is of a special form  $\inf_{f\in\mathcal{F}_{CNN}}R^\phi(f) -R^\phi(f^\phi_0)= \inf_{f\in\mathcal{F}_{CNN}}\mathbb{E}\vert f(X)-f^\phi_{0}(X)\vert^2,$ which leads to an improved exponent $n^{-2\alpha/(d+2\alpha)}$ rather than $n^{-2\alpha/(1.5d+4\alpha)}$ of the modified logistic or the exponential loss. Due to the low noise condition, the exponent of SVM risk bound $n^{-4\alpha(q+1)/\{3d+8\alpha(q+1)\}}$ is even more special, the worst case is $n^{-2\alpha/(1.5d+4\alpha)}$ when $q=0$ (high noise) and achieves the best possible rate $n^{-1/2}$ when $q=+\infty$ (no noise). Second the prefactors are also different for different loss functions. For least squares, its prefactor depends on the dimension $d$ by $O(d^{\alpha+1})$; for modified logistic, it is $O(d^{\alpha+1})$; for exponential loss (even modified), the dependence could be exponential type $O(d^{\alpha/2+1}\exp(d^{\alpha/2}))$; and for SVM the dependence is at most $O(d)$. These differences mainly come from the differences of Lipschitz constants of these loss functions on bounded intervals. On bounded interval $[-\mathcal{B},\mathcal{B}]$, the Lipschitz constants is $\mathcal{B}^2$ for the least squares loss; for modified logistic and exponential loss, it is $\exp(\mathcal{B})$; and for the SVM, the prefactor of the generalization rate does not depends on $\mathcal{B}$, since the SVM loss function is piecewise constant.
}

\section{Related works}
\label{related}

Binary classification is a basic and important problem in machine learning and statistics.
There is a vast literature on this problem and many methods have been developed in the literature, including the logistic regression \citep{cox1958regression,  kleinbaum2002logistic},  the probit model \citep{amemiya1978estimation}, the decision tree \citep{safavian1991survey}, the support vector machine \citep{cortes1995support}, the $\psi$-learning \citep{shen2003psi, liu2006mpsi}, and the random forests \citep{breiman2001random}, among others.
 When the inputs are assumed to be uniformly distributed on the surface of a sphere,
  \cite{kalai2008agnostically} derived non-asymptotic bounds for efficient binary prediction with half spaces by minimizing the misclassification error rate directly.
 Under the assumption that  the conditional distribution of the label $Y$ given the input $X$ is monotonic in $w^\top X$ for some $w\in\mathbb{R}^d$,
 \cite{kalai2009isotron}  provided an numerical efficient algorithm for minimizing the misclassification error.
 {\color{black}The convergence properties of the excess risk of the ERM for the misclassification 0-1 loss have been studied extensively under various conditions \citep{tsybakov2004optimal, massart2006, bartlett2006empirical,blumer1989learnability,zhivotovskiy2018localization,ben2014sample,kaariainen2006active}.
}
{\color{black} Different types of hypothesis spaces have been used in these approaches, including the linear space, the reproducing kernel Hilbert space and  the class of tree-based models. However, these earlier works did not consider the excess risk of classification using CNNs.}
 We refer to \citet{boucheron2005theory} for an excellent review on the theoretical results
 for classification.

\subsection{Approximation power of CNNs}
There have been intensive efforts devoted to understanding the theoretical properties of deep neural networks in recent years.
Many remarkable results have been obtained concerning the approximation power of  deep neural networks for multivariate functions; some examples of important recent works
include \cite{
yarotsky2017error,
yarotsky2018optimal,
lu2017expressive,
raghu2017expressive,
shen2019nonlinear,
shen2019deep,
nakada2019adaptive,
chen2019efficient}.
{\color{black}
These works focused on the approximation power for
ReLU activated feedforward neural networks, but did not consider the special structures of CNNs.}

The approximation power of  CNNs 
 has been studied by several authors, see, for example, \citet{bao2014approximation,zhou2018understanding,oono2019approximation,zhou2020theory} and \citet{zhou2020universality}.	Universality of approximation by ReLU activated deep CNNs was established in \cite{zhou2020universality} for target functions restricted to Sobolev space $H^{\alpha}(\mathbb{R}^d)$ with $\alpha>2+d/2$, i.e. the target function $f$ and all its partial derivatives up to order $\alpha$ are square integrable on $\mathbb{R}^d$. The approximation error satisfies $\Vert f-f_{FCNN}\Vert_\infty\leq C\sqrt{\log(L)}L^{-1/2-1/d}$,  where $c$ is an absolute constant depending on the Sobolev norm of $f$ and $f_{FCNN}$ is a $L$-layer fully convolutional neural network. Approximation power of downsampled convolutional neural networks on a special class of functions (ridge function) was studied by \cite{zhou2020theory}, where the target function is assumed to be of the form $f(x)=g(\xi^\top x)$, $g$ is a univariate H\"older continuous function of order $\alpha$, $\xi\in\mathbb{R}^d$ satisfying $\Vert\xi\Vert_2\leq1$. The downsampling operators (the same as a pooling operator) are applied between layers, who takes input from the previous layer and outputs a subvector of the input vector. The approximate rate is shown to be $\Vert f-f_{DCNN}\Vert_\infty\leq cN^{-\alpha}$, where $f_{DCNN}$ is a downsampled convolutional neural network with uniform filter length $s=O(4N+6)$ and width $\mathcal{W}=O(N)$. Further, it was also proved that a downsampled CNN $f_{DCNN}$ can compute the same function as a MLP $f_{MLP}$ does, with the total number of free parameters of $f_{DCNN}$ being at most 8 times of that of the $f_{MLP}$. However, this cannot fully explain the approximation advantage of CNN over MLP.

	Note that the CNNs defined in these works are different from ours in terms of the convolutional operation.
In the CNNs considered by \citet{zhou2018understanding,fang2020theory,zhou2020theory} and \citet{zhou2020universality}, each convolutional layer has only one filter with size $s$ and the induced weight matrix is of a special Toeplitz type with shape $(d_{in}+s)\times d_{in}$ and no fully-connected layer is allowed in hidden layers. Such a framework may not be general enough to cover the CNNs that have been successfully applied in modern machine learning tasks.

For general CNNs as considered in this work, \citet{bao2014approximation} studied the approximation power of the composited CNN $f_{CNN}=H\circ f_{FCNN}$, where $H$ is a fully-connected layer and $f_{FCNN}$ is a $L$-layer fully convolutional neural network. They proved that the composited CNN $f_{CNN}$ has similar approximation power as an MLP for a certain class of functions in the sense that, to achieve approximation accuracy $\epsilon$, the numbers of parameters needed for $f_{CNN}$ and MLP
have the same order in terms of  $\epsilon$.  \citet{oono2019approximation}  showed the approximation power of ResNet-type convolutional neural networks are at least as good as MLPs with its number of parameters having the same order.

\subsection{Error bounds in regression and classification}
Several stimulating papers have studied the statistical convergence properties of least squares estimators using feedforward neural networks in the context of nonparametric regression \citep{bauer2019deep, schmidt2020nonparametric,chen2019nonparametric,
kohler2019estimation, nakada2019adaptive, farrell2021deep, jiao2021dnr}.
They showed that nonparametric regression using feedforward neural networks with
a polynomial-growing network width $\mathcal{W}=O(d^\beta)$ achieves the optimal rate of convergence \citep{stone1982optimal} up to a $\log n$ factor,  however, with a prefactor
$C_d=O(a^d)$ for some $a \ge 2$ that grows exponentially with  the dimensionality $d$,
unless the network width $\mathcal{W}=O(a^d)$ and size $\mathcal{S}=O(a^d)$ grow exponentially as $d$ grows. For a general review on generalization and uniform convergence of deep learning models, see \citet{bartlett2021deep}.

When the inputs are assumed to be uniformly distributed on the surface of a sphere,
\citet{kalai2008agnostically} derived non-asymptotic bounds for efficient binary prediction with half spaces by minimizing the misclassification error rate directly.
\citet{kim2021fast}
 studied  the excess risk of empirical risk minimizer for classification under the hinge loss (SVM) using deep neural networks.
They aimed to establish the convergence rate under the Tsybakov noise condition \citep{mammen1999smooth,tsybakov2004optimal} in three different cases: smooth decision boundary, smooth conditional class probability $\eta$ and margin conditions.
{\color{black} However, \citet{kim2021fast} focused
on the use of deep feedforward neural networks, instead of deep CNNs, and placed restrictions on the shapes of the networks where the network is only allowed to have $O(\log n)$ depth and $O(n^\nu)$ width for some $\nu>0$. Their convergence rate results did not clearly describe the prefactors in the error bounds and they also did not address the conditions for circumventing the curse of dimensionality.}

{\color{black}
There are numerous studies on the convergence rate of excess misclassification 0-1 risk for binary classifications, see, e.g., \cite{vapnik1974theory,talagrand1994sharper,boucheron2005theory,tsybakov2004optimal,massart2006,bartlett2006empirical}. For the well-specified case assuming that the Bayes classifier $h_0$ belongs to the hypothesis class $\mathcal{H}$ with a finite VC dimension $d_{VC}$,  the convergence rate of  excess risk (stochastic error) for ERM can achieve the best possible $O(\sqrt{d_{VC}/n})$ without any specific assumptions on $\mathcal{H}$ or on the distribution of the data \citep{vapnik1974theory,talagrand1994sharper,boucheron2005theory}. If one further assumes the relationship between the predictor and  the label is deterministic, \cite{vapnik1974theory,blumer1989learnability} showed that the optimal rate is $O(d_{VC}/n)$ up to a logarithmic factor and it can be achieved by the ERM. Later, for the well-specified case, \cite{massart2006} proposed the bounded noise or Massart’s margin condition where $\vert 2\eta-1\vert$ is bounded away from 0, i.e $\vert 2\eta-1\vert\ge \xi$ for some $\xi>0$, and it is shown that the expected excess risk (stochastic error) of ERM is upper bounded by $O(d_{VC}\log(n\xi^2/d_{VC})/(n\xi))$ and the lower bound can be found in \cite{massart2006,zhivotovskiy2018localization}.  Also, for the well-specified case, \cite{mammen1999smooth,tsybakov2004optimal}  proposed the Tsybakov noise condition, i.e.,  there exist $c_{\text{noise}}>0$ and $q\in[0,\infty]$ such that for any $t>0$, $\mathbb{P}(\vert 2\eta(X)-1\vert\leq t)\leq c_{\text{noise}}t^q.$  Under Tsybakov noise condition, \cite{tsybakov2004optimal} showed that the minimax lower bound is $O(n^{-\beta(q+1)/\{\beta(q+2)+(d-1)(q+1)\}})$ over all smooth Bayes classifiers,  where their decision boundaries are generated by finite many $\beta$-H\"older smooth functions.
Still under the Tsybakov noise condition, \cite{audibert2007fast} showed that the minimax lower bound of the excess risk is $O(n^{-\alpha(q+1)/\{\alpha(q+2)+d\}})$ when $\eta$ belongs to the class of $\alpha$-H\"older smooth functions. For the agnostic or misspecified case that the Bayes classifier $h_0$ does not belong to the hypothesis class $\mathcal{H}$, \cite{ben2014sample} further assumed a deterministic scenario and proved a faster excess rate $O(d_{VC}/n)$ up to logarithm factors under the condition of the finite combinatorial diameter $D$ on $\mathcal{H}$ ($D(\mathcal{H})=\sup_{h,h^\prime\in\mathcal{H}}\vert\{x:h(x)\not=h^\prime(x)\}\vert$). But, in this case the ERM may suffer a slow rate and thus is not optimal. \cite{kaariainen2006active} later showed that for the misspecified case, as long as the  diameter $D$ is infinite, any estimator (including ERM) may have the slow rate $O(1/\sqrt{n})$. \cite{bousquet2019fast} considered a specific learning algorithm with a reject option, namely Chow’s reject option model, and they showed that it can achieve a faster rate $O(D/(n\xi))$ up to logarithm factors with the bounded noise condition $\xi$ and finite class diameter $D$ of $\mathcal{H}$.

For the relationship between the error bounds of regression and binary classification, \cite{yang1999minimax} studied the gap between minimax estimation of the conditional probability $\eta$ and the direct minimax estimation
of the Bayes classifier $h_0$. For any estimator $\hat{\eta}$ of the true conditional probability, one can use the ``plug-in" rules to induce a classifier $\hat{h}={\rm sign}(2\hat{\eta}-1)$, and an upper bound on the $L_2$ risk for estimating $\eta$ gives an upper bound on the excess misclassification risk for classification(estimating $h_0$). He showed that the two problems are in fact of the same difficulty in terms of rates of convergence under a sufficient condition, which is satisfied when $\eta$ belongs to some function classes,  including Besov (Sobolev), Lipschitz, and bounded variation.
\cite{marron1983optimal} considered the multiclassification problem with known prior probabilities. This work assumed the unknown underlying densities corresponding to the various classes satisfy some smoothness condition, and demonstrated that the optimal minimax rates of multiclassification excess risk is the same as that of the nonparametric regression in \cite{stone1982optimal}, which is $n^{-2\alpha/(d+2\alpha)}$ where $\alpha$ is the smoothness index.
}

To the best of our knowledge, there have not been systematic studies on the non-asymptotic excess risk bounds for classification with a general surrogate convex loss using CNNs. Our work contributes to the understanding of this deep learning method and filling the gap on this important topic.

\section{Conclusion}
\label{conclusion}
In this work, we establish non-asymptotic excess risk bounds for a class of classification methods using
CNNs. Our results are derived for a general class of convex surrogate losses and target functions with different modulus of continuity and are applicable to many widely used classification methods in practical machine learning tasks, including those with the least squares, the logistic, the exponential and the SVM hinge losses.
An important feature of our results is that we clearly define the prefactors in terms of the input data dimension
$d$ and other model parameters.  Our obtained excess risk bounds
 significantly improve over the currently available ones in the sense that,
 the prefactors of our obtained bounds depend
 on the dimensionality $d$ polynomially in some important models,  including the logistic regression and SVM.
 However, the prefactor of the excess risk bound with the exponential loss
 depends on $d$ exponentially, similar to the results
  in the existing works on nonparametric regression using deep neural networks.
 It is important to clearly describe the prefactors since they may dominant the error bounds
 in high-dimensional problems with a large $d$. We also show that the classification methods with CNNs can circumvent the curse of dimensionality if the input data is supported on a neighborhood of a low-dimensional manifold embedded in the ambient space $\mathbb{R}^d$.	

To establish the non-asymptotic excess risk bounds, we need to develop an
upper bound for the covering number for the class of general convolutional neural networks with a bias term in each convolutional layer, and derive new results on the approximation power of CNNs for any uniformly-continuous target functions. These results provide further insights into the complexity and the approximation power of general convolutional neural networks, which are of independent interest and may have applications in other problems involving CNNs.

Some important future problems include generalizing the theoretical results and techniques developed in this work to other types of classification methods such as the distance weighted discrimination \citep{marron2007, wang2016} and unsupervised learning problems that use CNNs for data representation and function approximation.



\acks{The work of Y. Jiao is supported in part by the National Science Foundation of China grant 11871474 and by the research fund of KLATASDSMOE of China.
The work of Y. Lin is supported by the Hong Kong Research Grants Council (Grant No.
14306219 and 14306620), the National Natural Science Foundation of China (Grant No.
11961028) and Direct Grants for Research, The Chinese University of Hong Kong.
The work of J. Huang is partially supported by the U.S. NSF grant DMS-1916199.}


\appendix

\section*{Appendix}
This appendix contains additional technical details and the proofs of the lemmas and theorems stated
in the paper.

\section{Covering number of CNNs} \label{appendixa}
	In this section, to give upper bounds of the estimation error, we study the covering number bounds for the class of functions implemented by CNNs. Firstly, the definitions of covering number and packing number are given below.
	\begin{definition}[Covering number]
		Let $(K,\Vert\cdot\Vert)$ be a metric space, let $C$ be a subset of $K$, and let $\epsilon$ be a positive real number. Let $B_\epsilon(x)$ denote the ball of radius $\epsilon$ centered at $x$. Then $C$ is called a $\epsilon$-covering of $K$, if $K\subset\cup_{x\in C}B_\epsilon(x).$ The covering number of the metric space $(K,\Vert\cdot\Vert)$ with any radius $\epsilon>0$ is the minimum cardinality of any $\epsilon$-covering, which is defined by $\mathcal{N}(K,\epsilon,\Vert\cdot\Vert)=\min\{\vert C\vert:C {\rm\ is\ a\ } \epsilon{\rm -covering\ of\ } K\}$.
	\end{definition}
	
	\begin{definition}[Packing number]
		Let $(K,\Vert\cdot\Vert)$ be a metric space, let $P$ be a subset of $K$, and let $\epsilon$ be a positive real number. Let $B_\epsilon(x)$ denote the ball of radius $\epsilon$ centered at $x$. Then $P$ is called a $\epsilon$-packing of $K$, if $\{B_\epsilon(x)\}_{x\in P}$ is pairwise disjoint. The $\epsilon$-packing number of the metric space $(K,\Vert\cdot\Vert)$ with any radius $\epsilon>0$ is the maximum cardinality of any $\epsilon$-packing, which is defined by $\mathcal{M}(K,\epsilon,\Vert\cdot\Vert)=\max\{\vert P\vert:P {\rm\ is\ a\ } \epsilon{\rm -packing\ of\ } K\}$.
	\end{definition}
	
	\begin{lemma}
		Let $(K,\Vert\cdot\Vert)$ be a metric space, and for any $\epsilon>0$, let $\mathcal{N}(K,\epsilon,\Vert\cdot\Vert)$ and $\mathcal{M}(K,\epsilon,\Vert\cdot\Vert)$ denote the $\epsilon$-covering number and $\epsilon$-packing number respectively, then
		$$\mathcal{M}(K,2\epsilon,\Vert\cdot\Vert)\leq\mathcal{N}(K,\epsilon,\Vert\cdot\Vert)\leq\mathcal{M}(K,\epsilon,\Vert\cdot\Vert).$$
		\begin{proof}
			For simplicity, we write $\mathcal{N}_\epsilon=\mathcal{N}(K,\epsilon,\Vert\cdot\Vert)$ and $\mathcal{M}_\epsilon=\mathcal{M}(K,\epsilon,\Vert\cdot\Vert)$.
			We firstly proof $\mathcal{M}_{2\epsilon}\leq\mathcal{N}_\epsilon$  by contradiction. Let $P=\{p_1,\ldots,p_{\mathcal{M}_{2\epsilon}}\}$ be any maximal $2\epsilon$-packing of $K$ and $C=\{c_1,\ldots,c_{\mathcal{N}_{\epsilon}}\}$ be any minimal $\epsilon$-covering of $K$. If $\mathcal{M}_{2\epsilon}\geq\mathcal{N}_\epsilon+1$, then we must have $p_i$ and $p_j$ belonging
			to the same $\epsilon$-ball $B_\epsilon(c_k)$ for some $i\not= j$ and $k$. This means that the distance between $p_i$ and $p_j$ cannot be more than the diameter of the ball, i.e. $\Vert p_i-p_j\Vert\leq2\epsilon$, which leads to a contradiction since $\Vert p_i-p_j\Vert>2\epsilon$ by the definition of packing.
			
			Secondly, we prove $\mathcal{N}_{\epsilon}\leq\mathcal{M}_\epsilon$ by showing that each maximal $\epsilon$-packing $P=\{p_1,\ldots,p_{\mathcal{M}_\epsilon}\}$ is also a  $\epsilon$-covering. Note that for any $x\in K\backslash P$, there exist a $p_i\in P$ such that $\Vert x-p_i\Vert\leq\epsilon$ (if this does not hold, then we can construct a bigger packing with
			$p_{\mathcal{M}_\epsilon+1}=x$). Thus $P$ is also a  $\epsilon$-covering and we have $\mathcal{N}_{\epsilon}\leq\mathcal{M}_\epsilon$ by the definition of covering.
		\end{proof}	
	\end{lemma}

	\begin{lemma}\label{a2}
		Let $a\geq0$ and $K=\{x:x\in\mathbb{R}^d,\Vert x\Vert_2\leq a\}$ where $\Vert\cdot\Vert_2$ is the $L_2$ norm of a vector, then for any $\epsilon>0$, the covering number of $K$ is bounded by
		$$\log\mathcal{N}(K,\epsilon,\Vert\cdot\Vert_2)\leq d\log(1+\frac{2a}{\epsilon}).$$
		
		\begin{proof}
			Let $P$ be a maximal $\epsilon$-packing of $K$ with packing number $\mathcal{M}_{\epsilon}$, then $\cup_{x\in P}B_{\epsilon/2}(x)\subset B_{a+\epsilon/2}(0)$ by the definition of packing. To compare the volume of these two sets, we have $\mathcal{M}_{\epsilon}(\epsilon/2)^d\leq(a+\epsilon/2)^d$, and thus
			$$\mathcal{N}(K,\epsilon,\Vert\cdot\Vert_2)\leq \mathcal{M}(K,\epsilon,\Vert\cdot\Vert_2)\leq (1+\frac{2a}{\epsilon})^d.$$
		\end{proof}
	\end{lemma}

	Define $\Vert\cdot\Vert_F$ as the Frobenius norm of a matrix, i.e. $\Vert A\Vert_F=\sqrt{\sum_{i=1}^{m}\sum_{j=1}^{n}a_{ij}^2}$ for a matrix $A=(a_{ij})\in\mathbb{R}^{m\times n}$, and define $\Vert\cdot\Vert_2$ as the spectral norm of a matrix , i.e. $\Vert A\Vert_2$ is the largest singular value of $A$ or the square root of the largest eigenvalue of the matrix $A^\top A$. For matrix $A=\in\mathbb{R}^{m\times n}$ of rank $r$, we have $\Vert A\Vert_2\leq\Vert A\Vert_F\leq\sqrt{r}\Vert A\Vert_2$, and $\Vert A\Vert_F=\Vert A\Vert_2$ when $A$ is a vector. Next, we give covering number upper bounds for single fully connected layer, single convolutional layer and general CNNs.
	\begin{lemma}[Covering number of a single fully connected layer]\label{cn-fnn}
		Let $W\in\mathbb{R}^{d_{out}\times d_{in}}$ and $b\in\mathbb{R}^{d_{out}}$ be the weight matrix and bias vector for a fully connected layer respectively, which satisfy $\Vert W\Vert_F+\Vert b\Vert_F\leq a$. Let $X\in\mathbb{R}^{d_{in}}$ be an input with bounded norm, i.e. $\Vert X\Vert_F=\Vert X\Vert_2<\infty$, and let $\mathcal{F}_{a}(X)=\{\sigma(WX+b)\in\mathbb{R}^{d_{out}}:\Vert W\Vert_F+\Vert b\Vert_F\leq a\}$, where $\sigma$ is the ReLU activation function or an activation function being $\kappa$-Lipschitz, then we have the following covering number bound
		$$\log \mathcal{N}(\mathcal{F}_{a}(X),\epsilon,\Vert\cdot\Vert_2)\leq (d_{in}+1)\times d_{out}\log(1+\frac{2a\kappa(\Vert X\Vert_F+1)}{\epsilon}).$$
		
		\begin{proof}
			Firstly note that for each $(W,b)$ in the product space $\mathbb{R}^{d_{out}\times d_{in}}\times \mathbb{R}^{d_{out}}$, the norm $\Vert(W,b)\Vert_F:=\Vert W\Vert_F+\Vert b\Vert_F$ is well-defined. 	
			Let $C=\{(W_1,b_1),\ldots,(W_N,b_N)\}$ be a minimal $\epsilon$-covering of $K=\{(W,b)\in\mathbb{R}^{d_{out}\times(d_{in}+1)}:\Vert W\Vert_F+\Vert b\Vert_F\leq a\}$ with covering number $N$. Then for any $(W,b)\in K$, there exist a $(W_i,b_i)$ such that $\Vert (W_i,b_i)-(W,b)\Vert_F\leq\epsilon$, and
			\begin{align*}
				&\Vert\sigma(W_iX+b_i)-\sigma(WX+b)\Vert_F\\
				\leq &\kappa\Vert(W_iX+b_i)-(WX+b)\Vert_F\\
				\leq &\kappa\Vert(W_i-W)X\Vert_F+\Vert(b_i-b)\Vert_F\\
				\leq&\kappa\Vert(W_i-W)\Vert_2 \Vert X\Vert_2+\rho\Vert(b_i-b)\Vert_2\\
				\leq&\epsilon\kappa(\Vert X\Vert_2+1).
			\end{align*}
			
			Thus $C_x=\{W_1X+b_1,\ldots,W_NX+b_N\}$ is a $\epsilon\kappa(\Vert X\Vert_2+1)$-covering of $\mathcal{F}_{a}(X)$. Note that $(W,b)$ can be reshaped into a $d_{out}\times (d_{in}+1)$-dimensional vector with its Euclidean norm $\sqrt{\Vert W\Vert_F^2+\Vert b\Vert_F^2}$ no greater than $\Vert W\Vert_F+\Vert b\Vert_F\leq a$, then by Lemma \ref{a2}, $\log \mathcal{N}(\mathcal{F}_{a}(X),\epsilon(\Vert X\Vert_2+1),\Vert\cdot\Vert_2)\leq N\leq (d_{in}+1)\times d_{out}\log(1+\frac{2a}{\epsilon}).$ Thus we have
			$$\log \mathcal{N}(\mathcal{F}_{a}(X),\epsilon,\Vert\cdot\Vert_2)\leq (d_{in}+1)\times d_{out}\log(1+\frac{2a\kappa(\Vert X\Vert_F+1)}{\epsilon}).$$
			
		\end{proof}
	\end{lemma}

	\begin{lemma}[Covering number of a single convolutional layer]\label{cn-cnn}
		Let $C=(c_1,\ldots,c_r)\in\mathbb{R}^{r\times s}$, $W^c\in\mathbb{R}^{d_{out}\times d_{in}}$ and $b^c\in\mathbb{R}^{d_{out}}$ be the convolutional matrix with $c$ filters with size $s$, induced weight matrix and bias vector for convolutional layer respectively, which satisfy $\Vert C\Vert_F+\Vert b^c\Vert_F\leq a$. Suppose each filter performs $m$ times operation, i.e. $d_{out}=r\times m$. Let $X\in\mathbb{R}^{d_{in}}$ be an input with bounded norm, i.e. $\Vert X\Vert_F=\Vert X\Vert_2<\infty$, and let $\mathcal{F}_{a}(X)=\{\sigma(W^cX+b^c)\in\mathbb{R}^{d_{out}}:\Vert C\Vert_F+\Vert b^c\Vert_F\leq a\}$, where $\sigma$ is the ReLU activation function or max-pooling operator being $\kappa$-Lipschitz, then we have the following covering number bound
		$$\log \mathcal{N}(\mathcal{F}_{a}(X),\epsilon,\Vert\cdot\Vert_2)\leq r\times(s+m)\log(1+\frac{2a\kappa(\sqrt{m}\Vert X\Vert_F+1)}{\epsilon}).$$
		
		\begin{proof}
			Firstly note that for each $(C,b^c)$ in the product space $\mathbb{R}^{c\times s}\times \mathbb{R}^{d_{out}}$, the norm $\Vert(C,b^c)\Vert=\Vert C\Vert_F+\Vert b^c\Vert_F$ is well-defined.
			Let $V=\{(C_1,b^c_1),\ldots,(C_N,b^c_N)\}$ be a minimal $\epsilon$-covering of $K=\{(C,b^c)\in\mathbb{R}^{c\times s}\times \mathbb{R}^{d_{out}}:\Vert C\Vert_F+\Vert b^c\Vert_F\leq a\}$ with covering number $N$. Then for any $(C,b^c)\in K$, there exist a $(C_i,b^c_i)$ such that $\Vert C_i-C\Vert_F +\Vert b^c_i-b^c\Vert_F\leq\epsilon$, and
			\begin{align*}
				&\Vert\sigma(W^c_iX+b^c_i)-\sigma(W^cX+b^c)\Vert_F\\
				\leq &\kappa\Vert(W^c_iX+b^c_i)-(W^cX+b^c)\Vert_F\\
				\leq &\kappa\Vert(W^c_i-W^c)X\Vert_F+\kappa\Vert(b^c_i-b^c)\Vert_F\\
				=	 & \kappa\sqrt{\sum_{i=1}^r\sum_{j=1}^m\Vert c_i^\top X_{S_j}\Vert_2^2}+\kappa\Vert(b^c_i-b^c)\Vert_F\\
				\leq & \kappa\Vert X\Vert_2\sqrt{\sum_{i=1}^r\sum_{j=1}^m\Vert c_i\Vert_2^2}+\kappa\Vert(b^c_i-b^c)\Vert_F\\
				= 	 & \kappa\sqrt{m}\Vert X\Vert_2\Vert C\Vert_F+\kappa\Vert(b^c_i-b^c)\Vert_F\\
				\leq & \kappa\epsilon(\sqrt{m}\Vert X\Vert_2+1).
			\end{align*}
			
			Thus $V_X=\{ W^c_1X+b^c_1,\ldots,W^c_NX+b^c_N\}$ is a $\kappa\epsilon(\sqrt{m}\Vert X\Vert_2+1)$-covering of $\mathcal{F}_{a}(X)$. Note that $(C,b^c)$ can be reshaped into a $rs+rm$-dimensional vector with its Euclidean norm equals $\Vert C\Vert_F+\Vert b^c\Vert_F\leq a$, then by Lemma \ref{a2}, $\log \mathcal{N}(\mathcal{F}_{a}(X),\kappa\epsilon(\sqrt{m}\Vert X\Vert_2+1),\Vert\cdot\Vert_F)\leq N\leq (rs+rm)\log(1+\frac{2a}{\epsilon}).$ Thus we have
			$$\log \mathcal{N}(\mathcal{F}_{a}(X),\epsilon,\Vert\cdot\Vert_2)\leq r\times(s+m) \log(1+\frac{2a(\sqrt{m}\Vert X\Vert_2+1)\kappa}{\epsilon}).$$
			
		\end{proof}
	\end{lemma}
	
\section{Proof of Theorems}
In this section, we prove Theorems  \ref{bound-cvnum-cnn}, \ref{bound-est}, \ref{bound-app},
\ref{gerrorbound},  \ref{gerrorbound-lowdim} and the SVM example.

\subsection{Proof of Theorem \ref{bound-cvnum-cnn}}
\begin{proof}
	The idea of our proof is mainly based on Lemma A.7 in \cite{bartlett2017spectrally}. We modified it by allowing the bias vector exists in each layer of the CNNs.
	
	Let $(\epsilon_1,\ldots,\epsilon_L)$ be given positive numbers, let $(\sigma_1,\ldots,\sigma_{L})$ be fixed Lipschitz mappings (where $\sigma_i$ is $\rho_i$-Lipschitz). Let $\mathcal{L}_F$ be a subset of indices $\{1,\ldots,L\}$ and $\mathcal{L}_C=\{1,\ldots,L\}\backslash\mathcal{L}_F$ be its complement, and let $A=(A_1,\ldots,A_L)$ be $L$ linear operators with their corresponding input dimension $(d_1,\ldots,d_L)$, where $A_l(X)=W_lX+b_l$ when $l\in\mathcal{L}_F$ with fully connected weights $W_l\in\mathbb{R}^{d_l\times d_{l+1}}$, bias $b_l\in\mathbb{R}^{d_{l+1}}$ and $A_l(X)=W^c_lX+b^c_l$ when $l\in\mathcal{L}_C$ with  $W^c_l\in\mathbb{R}^{d_l\times d_{l+1}}$ induced by convolutional weights $C_l\in\mathbb{R}^{r_i\times s_i}$ and bias $b^c_l\in\mathbb{R}^{d_{l+1}}$. Suppose each of the $r_i$ filter in $C_l$ performs $m_i$ operations. Given positive numbers $(a_1,\ldots,a_L)$, and sets $(\mathcal{A}_1,\ldots,\mathcal{A}_{L})$, where $\mathcal{A}_l=\{A_l:\Vert W_l\Vert_F+\Vert b_l\Vert_F\leq a_l\}$ for $l\in\mathcal{L}_F$ and $\mathcal{A}_l=\{ A_l:\Vert C_l\Vert_F+\Vert b^c_l\Vert_F\leq a_l\}$ for $l\in\mathcal{L}_C$, and assume the support of the input $\mathcal{X}\subset\mathbb{R}^{d_0}$ is a bounced set, i.e. $\sup_{X\in\mathcal{X}}\Vert X\Vert_2\leq B$ for $B\geq1$. Define the function class implemented by CNNs
	$$\mathcal{F}_{CNN}=\{f:\mathcal{X}\to\mathbb{R}^{d_L}:f= \sigma_{L}\circ A_L\cdots\circ\sigma_1\circ A_1, A_l\in\mathcal{A}_l {\ \rm for\ } l=1\ldots,L\},$$
	and for any $f_1,f_2\in\mathcal{F}$, define the $\Vert\cdot\Vert_\infty$ metric on $\mathcal{F}_{CNN}$ by $\sup_{X\in\mathcal{X}}\Vert f_1(X)-f_2(X)\Vert_\infty$. And let $\mathcal{B}:=\sup_{f\in\mathcal{F}_{CNN}}\Vert f\Vert_\infty$ denote the bound of functions in $\mathcal{F}_{CNN}$ and let $\mathcal{S}=\sum_{l\in\mathcal{L}_F}(d_{l-1}+1)\times d_{l}+\sum_{l\in\mathcal{L}_C}(s_i+m_i)\times r_{l}$ denote the number of parameters for networks in $\mathcal{F}_{CNN}.$
	Then for any $\epsilon>0$, we have
	$$\log\mathcal{N}(\mathcal{F}_{CNN},\epsilon,\Vert\cdot\Vert_\infty)\leq 2\mathcal{S}\Big(\frac{L\mathcal{B}}{\epsilon}\Big)^{1/2}.$$

We now inductively prove the result in four steps. For $l=1,\ldots,L$, define $\mathcal{F}_{l}=\{f:\mathcal{X}\to\mathbb{R}^{d_l}:f= \sigma_{l}\circ A_l\cdots\circ\sigma_1\circ A_1, A_i\in\mathcal{A}_i {\ \rm for\ } i=1\ldots,l\}$. Note that $\mathcal{F}_{CNN}=\mathcal{F}_L$.

		{\it Step 1.}  Choose a minimal $\epsilon_1$-covering $K_1$ of $\mathcal{F}_1$, thus
		$$\vert K_1\vert\leq\mathcal{N}(\mathcal{F}_1,\epsilon_1,\Vert\cdot\Vert_\infty)=:\mathcal{N}_1.$$
		
		{\it Step 2.} For each element $f\in K_{l-1}$, we can construct an minimal $\epsilon_l$-covering $K_{l}(f)$ of the set of functions $\{\sigma_l\circ{A}_{l}\circ f:A_{l}\in\mathcal{A}_{l}\}$. Here $f=\sigma_{l-1}\circ A_{l-1}\cdots\circ\sigma_1\circ A_1\in\mathcal{F}_{l-1}$ for some certain $(A_{1},\ldots,A_{l-1})\in(\mathcal{A}_1,\ldots,\mathcal{A}_{l-1})$, thus
		$$\vert K_l(f)\vert\leq\sup_{(A_{1},\ldots,A_{i}), \forall i\leq l-1, A_{i}\in\mathcal{A}_i}\mathcal{N}(\{\sigma_l\circ{A}_{l}\circ f:f\in\mathcal{F}_{l-1}, A_{l}\in\mathcal{A}_{l}\},\epsilon_l,\Vert\cdot\Vert_\infty):=\mathcal{N}_l.$$
		By the definition of $K_l=\cup_{f\in K_{l-1}}K_{l}(f)$, we have
		
		$$\vert K_l\vert \leq\vert K_{l-1}\vert \sup_{f\in K_{l-1}}\vert K_l(f)\vert\leq\mathcal{N}_l\Pi_{i=1}^{l-1}\mathcal{N}_i=\Pi_{i=1}^{l}\mathcal{N}_i.$$
		
		{\it Step 3.} We need to show such iterative constructed $K_L$ is indeed an $\epsilon$-covering of $\mathcal{F}_L=\mathcal{F}_{FNN}$ for some $\epsilon$ depending on $(\epsilon_1,\ldots,\epsilon_L)$, $(a_1,\ldots,a_L)$, $(\rho_1,\ldots,\rho_L)$ and $(d_0,\ldots,d_L)$. For any $f_L\in\mathcal{F}_L$, there exists $(A_1,\ldots,A_L)\in(\mathcal{A}_1,\ldots,\mathcal{A}_L)$ such that $f= \sigma_{L}\circ A_L\cdots\circ\sigma_1\circ A_1$, let $f_l=\sigma_{l}\circ A_l\cdots\circ\sigma_1\circ A_1$ for $l=1,\ldots,L-1$. By the definition of $K_1$, there exists $f_1^*\in K_1$ such that $\Vert f_1^*-f_1\Vert_\infty\leq \epsilon_1$. Define $\hat{f}^*_2=\sigma_2\circ A_2(f^*_1)$, then by the definition of $K_2$, there exists $f_2^*\in K_2$ such that $\Vert f_2^*-\sigma_2\circ A_2(f^*_1)\Vert_\infty\leq \epsilon_2$ and
		\begin{align*}
			\Vert f_2-f^*_2\Vert_\infty&\leq\Vert f_2-\sigma_2\circ A_2(f^*_1)\Vert_\infty+\Vert\sigma_2\circ A_2(f^*_1)-f_2^*\Vert_\infty \\
			&\leq\Vert \sigma_2\circ A_2(f_1)-\sigma_2\circ A_2(f^*_1)\Vert_\infty+\epsilon_2\\
			&\leq \rho_2a_2\Vert f_1^*-f_1\Vert_\infty+\epsilon_2\\
			&\leq \rho_2a_2\epsilon_1+\epsilon_2.
		\end{align*}
		By induction, there exists $f_L^*\in K_L$ such that
		$$\Vert f^*_L-f_L\Vert_\infty\leq\sum_{l}^{L}\epsilon_l\Pi_{i=l+1}^L\rho_ia_i.$$
		Let $\epsilon=\sum_{l=1}^{L}\epsilon_l\Pi_{i=l+1}^L\rho_ia_i$, by the result in Step 2, we know that $K_L$ is a $\epsilon$-covering of $\mathcal{F}_L=\mathcal{F}_{CNN}$ and
		$$\log\mathcal{N}(\mathcal{F}_{CNN},\epsilon,\Vert\cdot\Vert_\infty)\leq\log\vert K_L\vert\leq\sum_{i=1}^{L}\log\mathcal{N}_i.$$
		
		{\it Step 4.} We calculate $\log\mathcal{N}_i$ for each $i=1,\ldots,L$ and carefully choose $(\epsilon_1,\ldots,\epsilon_L)$ to give a covering number of $\mathcal{F}_{CNN}$ for any radius $\epsilon>0.$
		For $l\in\mathcal{L}_C$, we denote the number operations for each filter by $m_l=d_l/r_l$, and for $l\in\mathcal{L}_F$, we define $m_l=1$. Then for any $f_1\in\mathcal{F}_1$, we have $\Vert f_1\Vert_\infty\leq \rho_1a_1(\sqrt{m_1}B+1)$. By induction, it is easy to find for any $f_l\in\mathcal{F}_l$ and $l=1,\ldots,L$ we have $\Vert f_l\Vert_\infty\leq B\Pi_{i=1}^l\sqrt{m_i}\rho_ia_i+\sum_{i=1}^l\Pi_{j=i}^l\sqrt{m_j}\rho_ja_j:=B_l$.
		For $l\in\mathcal{L}_F$,  based on Lemma \ref{cn-fnn}, we have
		$$\log\mathcal{N}_l=\log \mathcal{N}(\{\sigma_l\circ A_l\circ f\in\mathcal{F}_{l-1},A_l\in\mathcal{A}_l\},\epsilon_l,\Vert\cdot\Vert_\infty)\leq (d_{l-1}+1)\times d_{l}\log(1+\frac{2\rho_la_l(\sqrt{m_l}B_{l-1}+1)}{\epsilon_l}),$$
		and for $l\in\mathcal{L}_C$,  based on Lemma \ref{cn-cnn},  we have
		$$\log\mathcal{N}_l=\log \mathcal{N}(\{\sigma_l\circ A_l\circ f\in\mathcal{F}_{l-1},A_l\in\mathcal{A}_l\},\epsilon_l,\Vert\cdot\Vert_\infty)\leq (s_i+m_i)\times r_{l}\log(1+\frac{2\rho_la_l(\sqrt{m_l}B_{l-1}+1)}{\epsilon_l}).$$
		Then, we let $\epsilon_l=e_l\epsilon/ \Pi_{i=l+1}^{L}\rho_ia_i$, and $e_l=(\sqrt{m_l}B_{l-1}+1)\Pi_{i=l}^L\rho_ia_i/\sum_{i=1}^L(\sqrt{m_i}B_{i-1}+1)\Pi_{j=i}^L\rho_ja_j$ which satisfies $\sum_{l=1}^Le_l=1$ and the requirement $\epsilon=\sum_{l=1}^{L}\epsilon_l\Pi_{i=l+1}^L\rho_ia_i$ in Step 3. By the result in Step 3,
		\begin{align*}
			&\log\mathcal{N}(\mathcal{F}_{CNN},\epsilon,\Vert\cdot\Vert_\infty)\leq\sum_{l\in\mathcal{L}_F}\log\mathcal{N}_l+\sum_{l\in\mathcal{L}_C}\log\mathcal{N}_l\\
			\leq& \sum_{l\in\mathcal{L}_F}(d_{l-1}+1)\times d_{l}\log(1+\frac{2\rho_la_l(\sqrt{m_l}B_{l-1}+1)}{\epsilon_l})+\sum_{l\in\mathcal{L}_C}(s_i+m_i)\times r_{l}\log(1+\frac{2\rho_la_l(\sqrt{m_l}B_{l-1}+1)}{\epsilon_l})\\
			=& \Big(\sum_{l\in\mathcal{L}_F}(d_{l-1}+1)\times d_{l}+\sum_{l\in\mathcal{L}_C}(s_i+m_i)\times r_{l}\Big)\log\Big(1+\frac{2\sum_{l=1}^L(\sqrt{m_l}B_{l-1}+1)\Pi_{i=l}^L\rho_ia_i)}{\epsilon}\Big)\\
			\leq & \Big(\sum_{l\in\mathcal{L}_F}(d_{l-1}+1)\times d_{l}+\sum_{l\in\mathcal{L}_C}(s_i+m_i)\times r_{l}\Big)\log\Big(1+\frac{4\sum_{l=1}^L\sqrt{m_l}B_{l-1}\Pi_{i=l}^L\rho_ia_i}{\epsilon}\Big)\\
			\leq & \Big(\sum_{l\in\mathcal{L}_F}(d_{l-1}+1)\times d_{l}+\sum_{l\in\mathcal{L}_C}(s_i+m_i)\times r_{l}\Big)\log\Big(1+\frac{4LB_L}{\epsilon}\Big).
		\end{align*}
		The first inequality follows form Lemma \ref{cn-fnn} and \ref{cn-cnn}.  The second last inequality follows form $\sqrt{m_l}B_{l-1}\geq1$ and the last inequality follows form  $\sqrt{m_l}B_{l-1}\Pi_{i=l}^L\rho_ia_i\leq B_L$.
		If we let $\mathcal{S}=\sum_{l\in\mathcal{L}_F}(d_{l-1}+1)\times d_{l}+\sum_{l\in\mathcal{L}_C}(s_i+m_i)\times r_{l}$ denote the number of parameters in the CNN, let $\mathcal{B}=B_L$ denote the bound of functions in $\mathcal{F}_{CNN}$, then
		$$\log\mathcal{N}(\mathcal{F}_{CNN},\epsilon,\Vert\cdot\Vert_\infty)\leq\mathcal{S}\log\Big(1+\frac{4\mathcal{B}L}{\epsilon}\Big)\leq2\mathcal{S}\Big( \frac{4\mathcal{B}L}{\epsilon}\Big)^{1/2},$$
		where the second inequality follows form $\sqrt{x}\geq\log(1+x)$ for $x\geq0$.
	\end{proof}
	
	\subsection{Proof of Theorem \ref{bound-est}}
\begin{proof}
	By Lemma \ref{lemma2}, for any loss $\phi$ and random sample $S=\{(X_i,Y_i)\}_{i=1}^n$,  the excess $\phi$-risk of ERM defined in (\ref{germ}) satisfies
	\begin{align*}
		R^\phi(\hat{f}_n^\phi)-R^\phi(f^\phi_0)\leq&  2\sup_{f\in\mathcal{F}}\vert R^\phi(f)-R^\phi_n(f)\vert+\inf_{f\in\mathcal{F}}R^\phi(f) -R^\phi(f^\phi_0),
	\end{align*}
	where $R^\phi_n(f)$ is the empirical risk of $f$ on sample $S$ under loss $\phi$ which is defined by $R^\phi_n=\sum_{i=1}^{n}\phi(Y_if(X_i))/n$. Next, we take care the first term $2\sup_{f\in\mathcal{F_{CNN}}}\vert R^\phi(f)-R^\phi_n(f)\vert$ here when the function class is taken as functions implemented by CNNs $\mathcal{F}_{CNN}$. Our proof consists of two main Steps.
	
	{\it Step 1.} For any given $\delta>0$, with probability at least $1-\delta$, the following holds,
	
	$$\sup_{f\in\mathcal{F_{CNN}}}\vert R^\phi(f)-R^\phi_n(f)\vert\leq 2B^\phi\mathcal{R}_n(\mathcal{F}_{CNN})+B_\phi\mathcal{B}\sqrt{\frac{2\log(1/\delta)}{n}},$$
	where $\mathcal{R}_S(\mathcal{F}_{CNN})=\mathbb{E}_{\sigma}\Big[\sup_{f\in\mathcal{F}_{CNN}}\frac{1}{n}\sum_{i=1}^n\sigma_if(X_i)\Big]$ is the empirical Rademacher complexity of $\mathcal{F}_{CNN}$ with respect to the sample $S$ and $\sigma=(\sigma_1,\ldots,\sigma_n)^\top$, with $\sigma_i$ being independent Rademacher random variables, i.e. uniform random variables taking values in $\{+1,-1\}$. We let $\mathcal{R}_n(\cdot)=\mathbb{E}\mathcal{R}_S(\cdot)$ denote the expectation of the empirical Rademacher complexity.
	Define $$G^\phi(S)=\sup_{f\in\mathcal{F}_{CNN}}\Big[\frac{1}{n}\sum_{i=1}^{n}\{\phi(Y_if(X_i))-\mathbb{E}\phi(Y_if(X_i))\}\Big],$$
	and let $S^\prime$ be another sample differing form $S$ by exactly one point, say $(X_1,Y_1)$ in $S$ and $(X^\prime_1,Y^\prime_1)$ in $S^\prime$. Then, by the definition of suprema and Assumption \ref{assumption1}, we have
	\begin{align*}
		G^\phi(S^\prime)-G^\phi(S)&\leq \sup_{f\in\mathcal{F}_{CNN}}\frac{1}{n}\sum_{i=1}^{n}\{\phi(Y_if(X_i))-\phi(Y^\prime_if(X^\prime_i))\}\\
		&=\sup_{f\in\mathcal{F}_{CNN}}\frac{1}{n}\{\phi(Y_1f(X_1))-\phi(Y^\prime_1f(X^\prime_1))\}\\
		&\leq B_\phi\vert Y_1f(X_1)-Y^\prime_1f(X^\prime_1)\vert/n\\
		&\leq \frac{2\mathcal{B}B_\phi}{n}.
	\end{align*}
	Then, by McDiarmid’s inequality, for any $\delta>0$, with probability at least $1-\delta$, it follows
	\begin{equation}\label{eq1}
		G^\phi(S)\leq\mathbb{E}_S\{G^\phi(S)\}+\mathcal{B}B_\phi\sqrt{\frac{2\log(1/\delta)}{n}}.
	\end{equation}
	
	Now let $S^\prime$ be another sample independent with $S$ and $\sigma=(\sigma_1,\ldots,\sigma_n)^\top$ be independent Rademacher random variables. We next bound the expectation of $G^\phi(S)$ as below,
	\begin{align*}
		\mathbb{E}_S\{G^\phi(S)\}&=\mathbb{E}_S\Bigg(\sup_{f\in\mathcal{F}_{CNN}}\Big[\frac{1}{n}\sum_{i=1}^{n}\{\phi(Y_if(X_i))-\mathbb{E}\phi(Y_if(X_i))\}\Big]\Bigg)\\
		&=\mathbb{E}_S\Bigg(\sup_{f\in\mathcal{F}_{CNN}}\mathbb{E}_{S^\prime}\Big[\frac{1}{n}\sum_{i=1}^{n}\{\phi(Y_if(X_i))-\phi(Y^\prime_if(X^\prime_i))\}\Big]\Bigg)\\
		&\leq\mathbb{E}_{S,S^\prime}\Bigg(\sup_{f\in\mathcal{F}_{CNN}}\Big[\frac{1}{n}\sum_{i=1}^{n}\{\phi(Y_if(X_i))-\phi(Y^\prime_if(X^\prime_i))\}\Big]\Bigg)\\
		&=\mathbb{E}_{S,S^\prime,\sigma}\Bigg(\sup_{f\in\mathcal{F}_{CNN}}\Big[\frac{1}{n}\sum_{i=1}^{n}\sigma_i\{\phi(Y_if(X_i))-\phi(Y^\prime_if(X^\prime_i))\}\Big]\Bigg)\\
		&\leq\mathbb{E}_{S,\sigma}\Bigg(\sup_{f\in\mathcal{F}_{CNN}}\Big[\frac{1}{n}\sum_{i=1}^{n}\sigma_i\phi(Y_if(X_i))\Big]\Bigg)+\mathbb{E}_{S^\prime,\sigma}\Bigg(\sup_{f\in\mathcal{F}_{CNN}}\Big[\frac{1}{n}\sum_{i=1}^{n}\sigma_i\phi(Y^\prime_if(X^\prime_i))\Big]\Bigg)\\
		&=2\mathcal{R}_n(\mathcal{F}^\phi_{CNN}),
	\end{align*}
	where $\mathcal{F}^\phi_{CNN}=\{ (x,y)\mapsto\phi(yf(x)):f\in\mathcal{F}_{CNN}\}$ is the function class introduced by a fixed function $\phi$ and $\mathcal{F}_{CNN}$. By Assumption \ref{assumption1}, $\phi$ is $B_\phi$-Lipschitz, then by Talagrand’s lemma, see e.g. Lemma 5.7 in \cite{mohri2018foundations}, we have
	\begin{equation}\label{eq2}
		\mathcal{R}_n(\mathcal{F}^\phi_{CNN})\leq B^\phi\mathcal{R}_n(\mathcal{F}_{CNN}).
	\end{equation}
	
	{\it Step 2.} We bound the Rademacher complexity $\mathcal{R}_n(\mathcal{F}_{CNN})$ in (\ref{eq2}) by the covering number through Dudley’s entropy integral. Let $\mathcal{N}(\mathcal{F}_{CNN},\epsilon,\Vert\cdot\Vert_\infty)$ denote the covering number of $\mathcal{F}_{CNN}$ with radius $\epsilon>0$ under Frobenius norm $\Vert\cdot\Vert_F$ defined in Appendix \ref{appendixa}. Here since the output of each $f\in\mathcal{F}_{CNN}$ is a scalar, thus we can also use metric $\Vert\cdot\Vert_\infty$ to define the covering number. By Dudley’s entropy integral, we have
	\begin{align*}
		\mathcal{R}_n(\mathcal{F}_{CNN})&\leq \inf_{a>0}\Big( {4a}+{12}\int_{a}^{\mathcal{B}}\sqrt{\frac{\log\mathcal{N}(\mathcal{F}_{CNN},\epsilon,\Vert\cdot\Vert_\infty)}{n}}d\epsilon \Big)\\
		&\leq \inf_{a>0}\Big( {4a}+{12}\int_{a}^{\mathcal{B}}\frac{(2\mathcal{S})^{1/2}(\mathcal{B}L)^{1/4}}{n^{1/2}\epsilon^{1/4}}d\epsilon \Big)\\
		&= \inf_{a>0}\Big( {4a}+\frac{16\sqrt{2}\mathcal{S}^{1/2}(\mathcal{B}L)^{1/4}}{n^{1/2}}(\mathcal{B}^{3/4}-a^{3/4}) \Big).
	\end{align*}
	Taking $a=4\cdot3^4\cdot\mathcal{S}^2\mathcal{B}L/n^2$, we have,
	\begin{align*}
		\mathcal{R}_n(\mathcal{F}_{CNN})&\leq \inf_{a>0}\Big( {4a}+\frac{16\sqrt{2}\mathcal{S}^{1/2}(\mathcal{B}L)^{1/4}}{n^{1/2}}(\mathcal{B}^{3/4}-a^{3/4}) \Big)\\
		&\leq \frac{16\cdot3^4\cdot\mathcal{S}^2\mathcal{B}L}{n^2}+\frac{16\sqrt{2}\mathcal{B}\mathcal{S}^{1/2}L^{1/4}}{n^{1/2}}-\frac{64\cdot3^3\cdot\mathcal{S}^2\mathcal{B}L}{n^2}\\
		&=\frac{16\sqrt{2}\mathcal{B}\mathcal{S}^{1/2}L^{1/4}}{n^{1/2}}-\frac{16\cdot3^3\cdot\mathcal{S}^2\mathcal{B}L}{n^2}\\
		&\leq \frac{16\sqrt{2}\mathcal{B}\mathcal{S}^{1/2}L^{1/4}}{n^{1/2}}.
	\end{align*}
	Combine above inequality, (\ref{eq1}) and (\ref{eq2}), we have for any $\delta>0$, with probability at least $1-\delta$,
	\begin{equation*}
		R^\phi(\hat{f}_n^\phi)-R^\phi(f^\phi_0)\leq \frac{16\sqrt{2}B_\phi\mathcal{B}\mathcal{S}^{1/2}L^{1/4}}{\sqrt{n}}+2B_\phi\mathcal{B}\sqrt{\frac{2\log(1/\delta)}{n}}+\inf_{f\in\mathcal{F}_{CNN}}R^\phi(f) -R^\phi(f^\phi_0).
	\end{equation*}
	\end{proof}
	
	\subsection{Proof of Theorem \ref{bound-app}}
\begin{proof}
	Firstly, we prove that under Assumption \ref{assumption1} and \ref{assumption2}, $f_{0,T}^\phi$,  the truncated version of the target function $f^\phi_0$, is bounded by $T$ and continuous on $[0,1]^d$. Recall that for any $a\in\mathbb{R}$ and $\eta \in[0,1]$, the conditional risk function $H^\phi(\eta,a)=\eta\phi(a)+(1-\eta)\phi(-a)$.
	Given $\eta\in[0,1]$, the function $dH^\phi(\eta,a)/da=\eta\phi^\prime(a)-(1-\eta)\phi^\prime(-a)$ is continuous in both $\eta$ and $a$. Thus the solution $a(\eta)=\arg\min_{a\in\bar{\mathbb{R}}}H(\eta,\alpha)$ is continuous with respect to $\eta$, and $\vert a(\eta)\vert<\infty$ when $\eta\in(0,1)$. By Assumption \ref{assumption2} (a), $\eta(x)$ is continuous on $\{x\in[0,1]^d:\eta(x)\leq1-\delta\}$ for any $\delta\in(0,1)$, thus $f_0^\phi(x)=f_0^\phi(\eta(x))$ is continuous on $\{x\in[0,1]^d:\vert f_0^\phi(\eta(x))\vert\leq T\}$ for any $T>0$. And for any $T>0$, the set $\{x\in[0,1]^d:\vert f_0^\phi(x)\vert\leq T\}$ is a compact set. Thus $f_{0,T}^\phi$, the truncated version of target function is continuous on $[0,1]^d$.
	
	Recall that for any $f$, the $\phi$-risk is defined by $R^\phi(f)=\mathbb{E}\phi(Yf(X))$, then for $f_{0,T}^\phi$, the truncated version of target function satisfies
	\begin{align*}
		R^\phi(f_{0,T}^\phi)-R^\phi(f_0^\phi)&=\mathbb{E}\{\phi(Yf_{0,T}^\phi(X))-\phi(Yf_0^\phi(X))\}\\
		&=\mathbb{E}\Big[\{\phi(Yf_{0,T}^\phi(X))-\phi(Yf_0^\phi(X))\}\{\mathbbm{1}(\vert f_0^\phi(X)\vert\leq B)+\mathbbm{1}(\vert f_0^\phi(X)\vert>B)\}\Big]\\
		&=\mathbb{E}\Big[\mathbbm{1}(\vert f_0^\phi(X)\vert>T)\{\phi(Yf_{0,T}^\phi(X))-\phi(Yf_0^\phi(X))\}\Big]\\
		&\leq \inf_{\vert a\vert\leq B}\phi(a)-\inf_{a\in\bar{\mathbb{R}}}\phi(a).
	\end{align*}
	Notice that,
	$$\inf_{f\in\mathcal{F}_{CNN}}R^\phi(f)-R^\phi(f_0^\phi)=	\inf_{f\in\mathcal{F}_{CNN}}R^\phi(f)-R^\phi(f_{0,T}^\phi)+R^\phi(f_{0,T}^\phi)-R^\phi(f_0^\phi),$$
	then left thing to do is to prove
	$$\inf_{f\in\mathcal{F}_{CNN}}R^\phi(f)-R^\phi(f_{0,T}^\phi)\leq B_\phi \inf_{f\in\mathcal{F}_{CNN}}\mathbb{E}\vert f(X)-f^\phi_{0,T}(X)\vert.$$
	With Assumption \ref{assumption1}, $\phi$ is $\mathcal{B}_\phi$-Lipschitz, and it is a straightforward proof that
	\begin{align*}
		\inf_{f\in\mathcal{F}_{CNN}}R^\phi(f)-R^\phi(f_{0,T}^\phi)&=\inf_{f\in\mathcal{F}_{CNN}}\mathbb{E}\{\phi(Yf(X))-\phi(Yf_{0,T}^\phi(X))\}\\
		&\leq \inf_{f\in\mathcal{F}_{CNN}}\mathbb{E}\{\mathcal{B}_\phi\vert Yf(X)-Yf_{0,T}^\phi(X)\vert\}\\
		&=B_\phi\inf_{f\in\mathcal{F}_{CNN}}\mathbb{E}\vert f(X)-f_{0,T}^\phi(X)\vert.
	\end{align*}
	\end{proof}
	
	\subsection{Proof of Theorem \ref{gerrorbound}}
\begin{proof}
	Let $K\in\mathbb{N}^+$ and $\tau\in(0,1/K)$, define a region $\Omega([0,1]^d,K,\tau)$ of $[0,1]^d$ as $$\Omega([0,1]^d,K,\varepsilon)=\cup_{i=1}^d\{x=[x_1,x_2,...,x_d]^T:x_i\in\cup_{k=1}^{K-1}(k/K-\tau,k/K)\}.$$
	By Theorem 2.1 of \cite{shen2019deep}, for any $M, N\in\mathbb{N}^+$, there exists a function $f^\phi_*\in\mathcal{F}_{MLP}$ with $f^\phi_{0,T}(\textbf{0})+\omega_{f^\phi_{0,T} }(\sqrt{d})\leq \mathcal{B}$, depth $L=12M+14$ and width $\mathcal{W}=\max\{4d\lfloor N^{\frac{1}{d}}\rfloor+3d,12N+8\}$ such that $\Vert f^\phi_*\Vert_\infty \leq
	\vert f^\phi_{0,T}(\textbf{0})\vert+\omega_{f^\phi_{0,T} }(\sqrt{d})$ and
	$$\vert f^\phi_*(x)-f^\phi_{0,T}(x)\vert\leq18\sqrt{d} \omega_{f^\phi_{0,T}}(N^{-2/d}M^{-2/d}),$$
	for any $x\in[0,1]^d\backslash \Omega([0,1]^d,K,\tau)$ where $K=\lfloor N^{1/d}\rfloor^2\lfloor M^{1/d}\rfloor^2$ and $\tau$ is an arbitrary number in $(0,\frac{1}{3K}]$. Note that the Lebesgue measure of $\Omega([0,1]^d,K,\tau)$ is no more than $dK\tau$ which can be arbitrarily small if $\tau$ is arbitrarily small. Since $P$ is absolutely continuous with respect to Lebesgue measure, then we have
	$$\mathbb{E}\Vert f^\phi_*(X)-f^\phi_{0,T}(X)\Vert_2\leq18\sqrt{d} \omega_{f^\phi_{0,T}}(N^{-2/d}M^{-2/d}).$$
	
	Besides, as the class of functions $\mathcal{F}_{MLP}$ is a subclass of $\mathcal{F}_{CNN}$ with layer $L$, width $\mathcal{W}$, bound $\mathcal{B}$ and size $\mathcal{S}$. Thus there exists a function $f$ implemented by CNN with its depth no more than $L=12M+14$ and its width no more than $\mathcal{W}=\max\{4d\lfloor N^{\frac{1}{d}}\rfloor+3d,12N+8\}$ such that
	$$\mathbb{E}\Vert f(X)-f^\phi_{0,T}(X)\Vert_2\leq18\sqrt{d} \omega_{f^\phi_{0,T}}(N^{-2/d}M^{-2/d}),$$
	and by Theorem \ref{bound-est} and \ref{bound-app}, for any $\delta>0$, with probability at least $1-\delta$,
	\begin{align*}
		R^\phi(\hat{f}_n)-R^\phi(f_0^\phi)\leq& \frac{16\sqrt{2}B_\phi\mathcal{B}\mathcal{S}^{1/2}L^{1/4}}{\sqrt{n}}+2B_\phi\mathcal{B}\sqrt{\frac{2\log(1/\delta)}{n}}\\
		+&18\sqrt{d}B_\phi \omega_{f^\phi_{0,T}}(N^{-2/d}M^{-2/d})+\inf_{\vert a\vert\leq T}\phi(a)-\inf_{a\in\bar{\mathbb{R}}}\phi(a).
	\end{align*}
	\end{proof}
\subsection{Proof of Lemma \ref{approx-lowDsupport}}
\begin{proof}
	According to Theorem 3.1 in \cite{baraniuk2009random}, there exists a linear projector $A\in\mathbb{R}^{d_\varepsilon\times d}$ that maps a low-dimensional manifold in a high-dimensional space to a low-dimensional space nearly preserving distance. To be exact, there exists a matrix $A\in\mathbb{R}^{d_\varepsilon\times d}$ such that
	$AA^T=(d/d_\varepsilon)I_{d_\varepsilon}$ where $I_{d_\varepsilon}$ is an identity matrix of size $d_\varepsilon\times d_\varepsilon$, and
	$$(1-\varepsilon)\vert x_1-x_2\vert\leq\vert Ax_1-Ax_2\vert\leq(1+\varepsilon)\vert x_1-x_2\vert,$$
	for any $x_1,x_2\in\mathcal{M}.$ And it is easy to check
	$$A(\mathcal{M}_\rho)\subseteq A([0,1]^d)\subseteq [-\sqrt{\frac{d}{d_\varepsilon}},\sqrt{\frac{d}{d_\varepsilon}}]^{d_\varepsilon}.$$
	For any $z\in A(\mathcal{M_\rho})$, define $x_z=\mathcal{SL}(\{x\in\mathcal{M}_\rho: Ax=z\})$ where $\mathcal{SL}(\cdot)$ is a set function which returns a unique element of a set. Note if $Ax=z$, then it is not necessary that $x=x_z$. For the high-dimensional function $f^\phi_{0,T}: \mathbb{R}^{d}\to\mathbb{R}^1$, we define its low-dimensional representation $\tilde{f}^\phi_{0,T}:\mathbb{R}^{d_\varepsilon}\to\mathbb{R}^1$ by
	$$\tilde{f}^\phi_{0,T}(z)=f^\phi_{0,T}(x_z), \quad {\rm for\ any} \ z\in A(\mathcal{M}_\rho)\subseteq\mathbb{R}^{d_\varepsilon}.$$
	For any $z_1,z_2\in A(\mathcal{M}_\rho)$, let $x_i=\mathcal{SL})(\{x\in\mathcal{M}_{\rho}, Ax=z_i\})$. By the definition of $\mathcal{M}_\rho$, there exist $\tilde{x}_1,\tilde{x}_2\in\mathcal{M}$ such that $\vert x_i-\tilde{x}_i\vert\leq\rho$ for $i=1,2$. Then
	\begin{align*}
		\vert \tilde{f}^\phi_{0,T}(z_1)-\tilde{f}^\phi_{0,T}(z_2)\vert&=\vert {f}^\phi_{0,T}(x_1)-{f}^\phi_{0,T}(x_2)\vert\leq\omega_{f^\phi_{0,T}}(\vert x_1-x_2\vert)\leq \omega_{f^\phi_{0,T}}(\vert \tilde{x}_1-\tilde{x}_2\vert+2\rho)\\
		&\leq \omega_{f^\phi_{0,T}}(\frac{1}{1-\varepsilon}\vert A\tilde{x}_1-A\tilde{x}_2\vert+2\rho)\\
		&\leq \omega_{f^\phi_{0,T}}(\frac{1}{1-\varepsilon}\vert A{x}_1-A{x}_2\vert+\frac{2\rho}{1-\varepsilon}\sqrt{\frac{d}{d_\varepsilon}}+2\rho)\\
		&\leq \omega_{f^\phi_{0,T}}(\frac{1}{1-\varepsilon}\vert z_1-z_2\vert+\frac{2\rho}{1-\varepsilon}\sqrt{\frac{d}{d_\varepsilon}}+2\rho).
	\end{align*}
	
	By Lemma 4.1 in \cite{shen2019deep}, there exists a $\tilde{g}$ defined on $\mathbb{R}^{d_\varepsilon}$,  whose  modulus of continuity is the same as $\tilde{f}^\phi_{0,T}$ such that
	$$\vert \tilde{g}(z)-\tilde{f}_{0,T}(z)\vert\leq\omega_{f^\phi_{0,T}}(\frac{2\rho}{1-\varepsilon}\sqrt{\frac{d}{d_\varepsilon}}+2\rho)$$
	for any $z\in A(\mathcal{M}_\rho)\subseteq\mathbb{R}^{d_\varepsilon}$. With $E=[-\sqrt{{d}/{d_\varepsilon}},\sqrt{{d}/{d_\varepsilon}}]^{d_\varepsilon}$, by Theorem 2.1 and 4.3 of \cite{shen2019deep}, for any $N,M\in\mathbb{N}_+$, there exists a function $\tilde{f}_{MLP}: \mathbb{R}^{d_\varepsilon}\to\mathbb{R}^1$ implemented by a ReLU MLP with width $\mathcal{W}=\max\{4d_\varepsilon\lfloor N^{1/d_\varepsilon}\rfloor+3d,12N+8 \}$ and depth $L=12M+14$ such that
	$$\vert \tilde{f}_{MLP}-\tilde{g}(z)\vert\leq18\sqrt{d_\varepsilon}\omega_{f^\phi_{0,T}}(N^{-2/d_\varepsilon}M^{-2/d_\varepsilon}),$$
	for any $z\in E\backslash \Omega(E)$ where $\Omega(E)$ is a subset of $E$ with arbitrarily small Lebesgue measure as well as $\Omega:=\{x\in\mathcal{M}_\rho: Ax\in\Omega(E)\}$ does.
	If we define $f_{MLP}=\tilde{f}_{MLP}\circ A$ which is $f_{MLP}(x)=\tilde{f}_{MLP}(Ax)$ for any $x\in[0,1]^d$, then $f_{MLP}\in\mathcal{F}_{MLP}$ is also a ReLU MLP with one more layer than $\tilde{f}_{MLP}$, where this additional layer computes linear transformation $A$.
	
	Besides, for any $x\in\mathcal{M}_\rho$, let $z=Ax$ and $x_z=\mathcal{SL}(\{x\in\mathcal{M}_\rho: Ax=z\})$. By the definition of $\mathcal{M}_\rho$, there exists $\bar{x},\bar{x}_z\in\mathcal{M}$ such that $\vert x-\bar{x}\vert\leq\rho$ and $\vert x_z-\bar{x}_z\vert\leq\rho$. Then we have
	\begin{align*}
		\vert x-x_z\vert&\leq\vert\bar{x}-\bar{x}_z\vert+2\rho\leq\vert A\bar{x}-A\bar{x}_z\vert/(1-\varepsilon)+2\rho\\
		&\leq(\vert A\bar{x}-A{x}\vert+\vert A{x}-A{x}_z\vert+\vert A{x}_z-A\bar{x}_z\vert)/(1-\varepsilon)+2\rho\\
		&\leq(\vert A\bar{x}-A{x}\vert+\vert A{x}_z-A\bar{x}_z\vert)/(1-\varepsilon)+2\rho\\
		&\leq 2\rho\{1+\sqrt{d/d_\varepsilon}/(1-\varepsilon)\}.
	\end{align*}
	Then for any $x\in\mathcal{M}_\rho\backslash\Omega$ and $z=Ax$,
	\begin{align*}
		\vert f_{MLP}(x)-f^\phi_{0,T}(x)\vert&\leq\vert f_{MLP}(x)-f^\phi_{0,T}(x_z)\vert+\vert f^\phi_{0,T}(x_z)-f^\phi_{0,T}(x)\vert\\
		&\leq\vert \tilde{f}_{MLP}(z)-\tilde{f}^\phi_{0,T}(z)\vert+\omega_{f^\phi_{0,T}}(\vert x-x_z\vert)\\
		&\leq \vert \tilde{f}_{MLP}(z)-\tilde{g}(z)\vert+\vert \tilde{g}(z)-\tilde{f}^\phi_{0,T}(z)\vert+\omega_{f^\phi_{0,T}}(2\rho\{1+\sqrt{d/d_\varepsilon}/(1-\varepsilon)\})\\
		&\leq 18\sqrt{d_\varepsilon}\omega_{f^\phi_{0,T}}(N^{-2/d_\varepsilon}M^{-2/d_\varepsilon})+2\omega_{f^\phi_{0,T}}(N^{-2/d_\varepsilon}M^{-2/d_\varepsilon})\\
		&\leq (18\sqrt{d_\varepsilon}+2)\omega_{f^\phi_{0,T}}(N^{-2/d_\varepsilon}M^{-2/d_\varepsilon}),
	\end{align*}
where the third and the fourth inequalities follow from $\rho\leq(1-\varepsilon)/\{2(\sqrt{d/d_\varepsilon}+1-\varepsilon)\}N^{-2/d_\varepsilon}M^{-2/d_\varepsilon}$. Since $P_X$ the probability measure of $X$ is absolutely continuous with respect to Lebesgue measure, then we have
	\begin{equation*}
		\mathbb{E}\vert f_{MLP}(X) -f^\phi_{0,T}(X)\vert \leq (18\sqrt{d_\varepsilon}+2)\omega_{f^\phi_{0,T}}(N^{-2/d_\varepsilon}M^{-2/d_\varepsilon}),
	\end{equation*}
	where $d_\varepsilon=O(d_\mathcal{M}\frac{\log(d/\varepsilon)}{\varepsilon^2})$ is assumed to satisfy $d_\varepsilon \ll d$.
\end{proof}
	\subsection{Proof of Theorem \ref{gerrorbound-lowdim}}
\begin{proof}
	The proof is similar to  that of Theorem \ref{gerrorbound}, except how we deal with $\inf_{f\in\mathcal{F}_{CNN}}\mathbb{E}\Vert f(X)-f^\phi_{0,T}(X)\Vert_2$.
Combining Theorems \ref{bound-est} and \ref{bound-app} and using Lemma \ref{approx-lowDsupport},  we have
	\begin{align*}
		R^\phi(\hat{f}_n^\phi)-R^\phi(f_0^\phi)\leq& \frac{16\sqrt{2}B_\phi\mathcal{B}\mathcal{S}^{1/2}L^{1/4}}{\sqrt{n}}+2B_\phi\mathcal{B}\sqrt{\frac{2\log(1/\delta)}{n}}\\
		+&(18\sqrt{d_\varepsilon}+2)B_\phi \omega_{f^\phi_{0,T}}(N^{-2/d_\varepsilon}M^{-2/d_\varepsilon})+\inf_{\vert a\vert\leq T}\phi(a)-\inf_{a\in\bar{\mathbb{R}}}\phi(a).
	\end{align*}
This completes the proof of Theorem \ref{gerrorbound-lowdim}.
	\end{proof}

	\subsection{Proof of the least squares example}
\begin{proof}
		For the least squares loss $\phi(a)=(1-a)^2$, we first prove that the approximation error is of a special form. For any $f\in\mathcal{F}_{CNN}$
		\begin{align*}
			R^\phi(f) -R^\phi(f^\phi_0)&=\mathbb{E}\{\phi(Yf(X))-\phi(Yf^\phi_0(X))\}\\
			&=\mathbb{E}\{(1-Yf(X))^2-(1-Yf^\phi_0(X))^2\}\\
			&=\mathbb{E}[f(X)^2-f^\phi_0(X)^2-2Y\{f(X)-f^\phi_0(X)\}]\\
			&=\mathbb{E}[(f(X)-f^\phi_0(X))\{f(X)+f^\phi_0(X)-2\mathbb{E}(Y\mid X)\}]\\
			&=\mathbb{E}[(f(X)-f^\phi_0(X))(f(X)+f^\phi_0(X)-2f^\phi_0(X))]\\
			&=\mathbb{E}\vert f(X)-f^\phi_0(X)\vert^2.
		\end{align*}
	Ans we have
		\begin{align*}
			\inf_{f\in\mathcal{F}_{CNN}}R^\phi(f) -R^\phi(f^\phi_0)&= \inf_{f\in\mathcal{F}_{CNN}}\mathbb{E}\vert f(X)-f^\phi_{0}(X)\vert^2.
		\end{align*}
	 Besides, note that for least squares $f^\phi_0=2\eta-1$ and $\Vert f^\phi_0\Vert_\infty=1$, thus taking $T=1$ in (\ref{fBphi}) is sufficient for the error control analysis. Let $\mathcal{F}_{CNN}$ be a class of CNNs  defined in (\ref{CNNs}) bounded by $\mathcal{B}\geq 1+\lambda d^{\alpha/2}$. By Theorem \ref{bound-app}, there exists a function $f$ implemented by CNN with its depth no more than $L=12M+14$ and its width no more than $\mathcal{W}=\max\{4d\lfloor N^{\frac{1}{d}}\rfloor+3d,12N+8\}$ such that
	$$\mathbb{E}\Vert f(X)-f^\phi_{0}(X)\Vert^2_2\leq18^2{d} \{\omega_{f^\phi_{0}}(N^{-2/d}M^{-2/d})\}^2,$$
	and	based on Theorem \ref{bound-est}, the empirical $\phi$-risk minimizer $\hat{f}^\phi_n$ defined in (\ref{erm-cnn}) satisfies, for any $\delta>0$, with probability at least $1-\delta$,
	\begin{equation*}
		R^\phi(\hat{f}^\phi_n)-R^\phi(f^\phi_0)\leq \frac{16\sqrt{2}B_\phi\mathcal{B}\mathcal{S}^{1/2}L^{1/4}}{\sqrt{n}}+2B_\phi\mathcal{B}\sqrt{\frac{2\log(1/\delta)}{n}}+2\times18^2\lambda{d} N^{-4\alpha/d}M^{-4\alpha/d},.
	\end{equation*}
	Let  $M=\lfloor n^{d/2(d+2\alpha)}\rfloor$, $N=1$, size (number of parameters) $\mathcal{S}\leq\mathcal{W}^2 L\leq \max\{49d^2,400\}\times\{\lfloor n^{d/2(d+2\alpha)}\rfloor+14\}$ and $\delta=\exp\{-n^{(d-2\alpha)/(d+2\alpha)}\}$. By plunging in $B_\phi=2\mathcal{B}$ and above values, we have with probability at least $1-\exp\{-n^{(d-2\alpha)/(d+2\alpha)}\}$,
	\begin{align*}
		R^\phi(\hat{f}^\phi_n)-R^\phi(f_0^\phi)\leq& \Big\{(32\sqrt{2}c+4\sqrt{2})\mathcal{B}^2+4\cdot18^2\lambda^2d(\max\{7d,20\})^{-\frac{4\alpha}{d}}\Big\}n^{-\frac{2\alpha}{(d+2\alpha)}},
	\end{align*}
	where $c=O(d)$ is a constant independent of $n,\mathcal{B},\alpha$ and $\lambda$.

The excess risk bound under the approximate low-dimensional manifold assumption can be obtained in a similar way.
	\end{proof}

	\subsection{Proof of the logistic loss example}
	
\begin{proof}
	For the modified logistic loss function $\phi(a)=\max\{\log\{1+\exp(-a)\},\tau\}$ with $\tau=\log\{1+\exp(-T)\}$, the corresponding measurable minimizer defined in (\ref{ermphi}) is $f^\phi_{0,T}$, the truncated version of the minimizer $f_0^\phi$ under the original logistic loss; see Table \ref{tab:2}. As a consequence, under the modified logistic loss, $\Delta_\phi(T)=0$ by definition since the infimum of the modified logistic loss can be achieved within $[-T,T]$.
	Let $\mathcal{F}_{CNN}$ be the class of CNNs  defined in (\ref{CNNs}) with $\min(\vert\log[\eta(\textbf{0})/\{1-\eta(\textbf{0})\}]\vert,T\}+\omega_{f^\phi_{0,T} }(\sqrt{d})\leq\mathcal{B}$, the number of layers $L\leq12\lfloor n^{2d/(3d+8\alpha)}\rfloor+14$, width $\mathcal{W}\leq\max\{7d,20\}$ and size $\mathcal{S}\leq\mathcal{W}^2 L\leq \max\{49d^2,400 \}\times(\lfloor n^{2d/(3d+8\alpha)}\rfloor+14)$.
	 Then by Theorem \ref{gerrorbound}, let $\delta=\exp\{-n^{d/(d+8\alpha/3)}\}$, we have with probability at least $1-\exp\{-n^{d/(d+8\alpha/3)}\}$, the excess $\phi$-risk of the ERM
	$\hat{f}^\phi_n$ defined in (\ref{erm-cnn}) satisfies
	\begin{align*}
	R^\phi(\hat{f}^\phi_n)-R^\phi(f_0^\phi)\leq& \Big\{(16\sqrt{2}c+2\sqrt{2})\mathcal{B}+72\lambda\exp(T)\sqrt{d} (\max\{7d,20\})^{-\frac{2\alpha}{d}}\Big\}n^{-\frac{\alpha}{0.75d+2\alpha}},
	\end{align*}
	where $c$ is a constant independent of $n,\mathcal{B},\alpha,\lambda$, and $c$ is independent of $d$ if $d\leq2$ otherwise $c=O(d)$.
	Alternatively,  (\ref{lgrisk}) can be simply written as 		
	$$R^\phi(\hat{f}^\phi_n)-R^\phi(f_0^\phi)\leq C(T,d)n^{-\frac{\alpha}{0.75d+2\alpha}}, $$
	where $C(T,d)=O(d(1+4\lambda d^{\alpha/2})\exp(T))$ is a constant independent of $n,\mathcal{B}$. Here the prefactor $C(T,d)$ depends on the dimension $d$ at a rate no more than $O(d^{\alpha/2+1})$.

The excess risk bound under the approximate low-dimensional manifold assumption can be obtained in a similar way.
\end{proof}
	
	\subsection{Proof of the exponential loss example}
\begin{proof}
	The proof is similar to that of the logistic loss example. We omitted the proof here.
\end{proof}

	\subsection{Proof of the SVM example}
\begin{proof}
	Based on Theorem \ref{bound-est}, our proof focus on controlling the approximation error $\inf_{f\in\mathcal{F}_{CNN}}R^\phi(f) -R^\phi(f^\phi_0)$. Recall that $f_0^\phi(x)={\rm sign}(2\eta(x)-1)$, we would firstly approximate H\"older continuous function $\eta$ by CNNs. Similar to the proof in Theorem \ref{gerrorbound}, for any $M, N\in\mathbb{N}^+$, there exists a function $\eta^*\in\mathcal{F}_{MLP}$ with bound $\mathcal{B}\ge1$, depth $L=12M+14$ and width $\mathcal{W}=\max\{4d\lfloor N^{\frac{1}{d}}\rfloor+3d,12N+8\}$ such that $\Vert \eta^*\Vert_\infty \leq
	\vert \eta(0)\vert+\omega_{\eta}(\sqrt{d})$ and
	$$\vert \eta^*(x)-\eta(x)\vert\leq18\sqrt{d} \omega_{\eta}(N^{-2/d}M^{-2/d}),$$
	for any $x\in[0,1]^d\backslash\Omega$ where $\Omega$ is a set with arbitrarily small Lebesgue measure. Let $\varepsilon_n=18\sqrt{d} \omega_{\eta}(N^{-2/d}M^{-2/d})$.
	
	We construct the neural network $f_{MLP}$ by adding one layer to $\eta^*$ as
	$$f_{MLP}(x)=2\sigma\Big[\frac{1}{\varepsilon_n}\{\eta^*(x)-\frac{1}{2}\}\Big]-2\sigma\Big[\frac{1}{\varepsilon_n}\{\eta^*(x)-\frac{1}{2}\}-1\Big]-1,$$
	where $\sigma$ is the ReLU activation function. Then we can see that $\Vert f_{MLP}\Vert_\infty\leq1$ and $f_{MLP}(x)=1$ if $\eta^*(x)\geq \varepsilon_n+1/2$, $f_{MLP}(x)=2(\eta^*(x)-1/2)/\varepsilon_n-1$ if $1/2+\varepsilon_n>\eta^*(x)\geq 1/2$ and $f_{MLP}(x)=-1$ if $\eta^*(x)<1/2.$ Let $\Omega_{\eta,\varepsilon}=\{x\in[0,1]^d:\vert2\eta(x)-1\vert>\varepsilon\}$ for $\epsilon>0$. Then for $x\in\Omega_{\eta,4\varepsilon_n}$, we have $\vert f_{MLP}(x)-f^\phi_0(x)\vert=0$, since $\eta^*(x)-1/2=(\eta(x)-1/2)-(\eta^*(x)-\eta(x))>\varepsilon_n$ when $\eta(x)-1/2>2\varepsilon_n$ and $\eta^*(x)-1/2<-\varepsilon_n$ when $\eta(x)-1/2<-2\varepsilon_n$. Then
	\begin{align*}
		R^\phi(f_{MLP}) -R^\phi(f^\phi_0)=&\mathbb{E}\Big[ \phi(Yf_{MLP}(X))-\phi(Yf_0^\phi(X))\Big]\\
		=&\mathbb{E}\Big[ \vert f_{MLP}(X)-f_0^\phi(X)\vert\vert2\eta(X)-1\vert\Big]\\
		=&\mathbb{E}_{X\in\Omega_{\eta,4\varepsilon_n}}\Big[ \vert f_{MLP}(X)-f_0^\phi(X)\vert\vert2\eta(X)-1\vert\Big]\\
		\leq&8\varepsilon_nP(\vert2\eta(X)-1\vert\leq4\varepsilon_n)\\
		\leq&8\times4^{q}\times c_{noise}\times\varepsilon_n^{q+1}.
	\end{align*}
	Setting the parameters in $\eta^*$ by $N=1$ and $M=12\lfloor n^{2d/\{3d+8\alpha(q+1)\}}\rfloor+14$ and $\mathcal{S}\leq\mathcal{W}^2 L\leq \max\{49d^2,400\}\times(\lfloor n^{2d/(3d+8\alpha(q+1))}\rfloor+14)$, and let $\delta$ be $\exp(-n^{3d/(3d+8\alpha(q+1))})$, then by Theorem \ref{bound-est}, with probability at least $1-\exp(-n^{3d/(3d+8\alpha(q+1))})$,
	\begin{align*}
		R^\phi(\hat{f}_n^\phi)-R^\phi(f_0^\phi)\leq& \Big\{16\sqrt{2}c+2\sqrt{2}+144\times4^q\lambda c_{noise}\sqrt{d} (\max\{7d,20\})^{-2\alpha/d}\Big\}\\
		&\hspace{4cm}\times n^{-4\alpha(q+1)/\{3d+8\alpha(q+1)\}},
	\end{align*}
	where $c$ is a constant independent with $n,\mathcal{B},\alpha,\lambda$ and $c$ is independent with $d$ if $d\leq2$ otherwise $c=O(d)$.

The excess risk bound under the approximate low-dimensional manifold assumption can be obtained in a similar way.
\end{proof}

\vskip 0.2in
\bibliography{cnn_jmlr}

\end{document}